\pdfoutput=1

\documentclass[11pt]{article}

\usepackage[preprint]{acl}

\usepackage{times}
\usepackage{latexsym}
\usepackage{amssymb}

\usepackage[T1]{fontenc}

\usepackage[utf8]{inputenc}

\usepackage{microtype}

\usepackage{inconsolata}

\usepackage{graphicx}

\usepackage{booktabs}
\usepackage{multirow}
\usepackage{caption}
\usepackage{makecell}
\usepackage{booktabs}
\usepackage{xcolor}
\usepackage[breakable]{tcolorbox}
\usepackage{enumitem}
\usepackage{float}

\usepackage{listings}
\usepackage{xcolor}

\usepackage{booktabs}
\usepackage{subcaption}

\usepackage{xcolor}
\usepackage{colortbl}
\usepackage{soul}
\usepackage{amsmath} 
\usepackage{longtable}
\usepackage{supertabular}
\usepackage[breakable]{tcolorbox}

\title{\textsc{MultiSessionCollab:} Learning User Preferences with Memory to Improve Long-Term Collaboration}

\author{
  Shuhaib Mehri \quad
  Priyanka Kargupta \quad
  Tal August \quad
  Dilek Hakkani-T\"ur \quad \\
  University of Illinois Urbana-Champaign \\
  \texttt{\{mehri2, pk36, taugust, dilek\}@illinois.edu}
}

\begin{document}
\maketitle

\begin{abstract}  
As conversational agents accumulate experience collaborating with users, adapting to user preferences is essential for fostering long-term relationships and improving collaboration quality over time. We introduce \textsc{MultiSessionCollab}, a benchmark that evaluates how well agents can learn user preferences and leverage them to improve collaboration quality throughout multiple sessions. To develop agents that succeed in this setting, we present long-term collaborative agents equipped with a memory that is specifically designed to learn user preferences across sessions and improve interactions. Moreover, we demonstrate that learning signals can be derived from user simulator behavior in \textsc{MultiSessionCollab} to train agents to generate more comprehensive reflections and update their memory more effectively. Extensive experiments show that equipping agents with our memory improves collaboration over time, yielding higher task success rates, more efficient interactions, and reduced user effort. Finally, we conduct a human user study that demonstrates that memory helps improve user experience in real-world settings.
\end{abstract}

\footnotetext[1]{  Code available at \url{https://github.com/Shuhaibm/multisessioncollab}}

\vspace{-0.5em}

\section{Introduction}

Throughout repeated interactions, humans naturally develop interpersonal relationships and adapt their behaviors to each other across various dimensions, including matching communication styles, establishing common ground, and refining mutual understanding \citep{burgoon1995interpersonal, niederhoffer2002linguistic, wilkes1992coordinating, clark1991grounding, clark1986referring}. This type of adaptation is particularly critical in human-AI collaboration, where users have interaction preferences (e.g., having key takeaways at the end of responses or receiving multiple options and their tradeoffs before making a decision) that influence how effectively they can understand and engage with agent responses \citep{sweller1988cognitive, shi2025models}. Rather than placing the cognitive burden on the user to repeatedly specify their preferences, conversational agents must be able to continuously learn and adhere to them over time \citep{bickmore2005establishing, shi2025models}.

\begin{figure*}[t]
    \centering
    \includegraphics[trim=0pt 0 0pt 0pt, clip, width=0.8\textwidth]{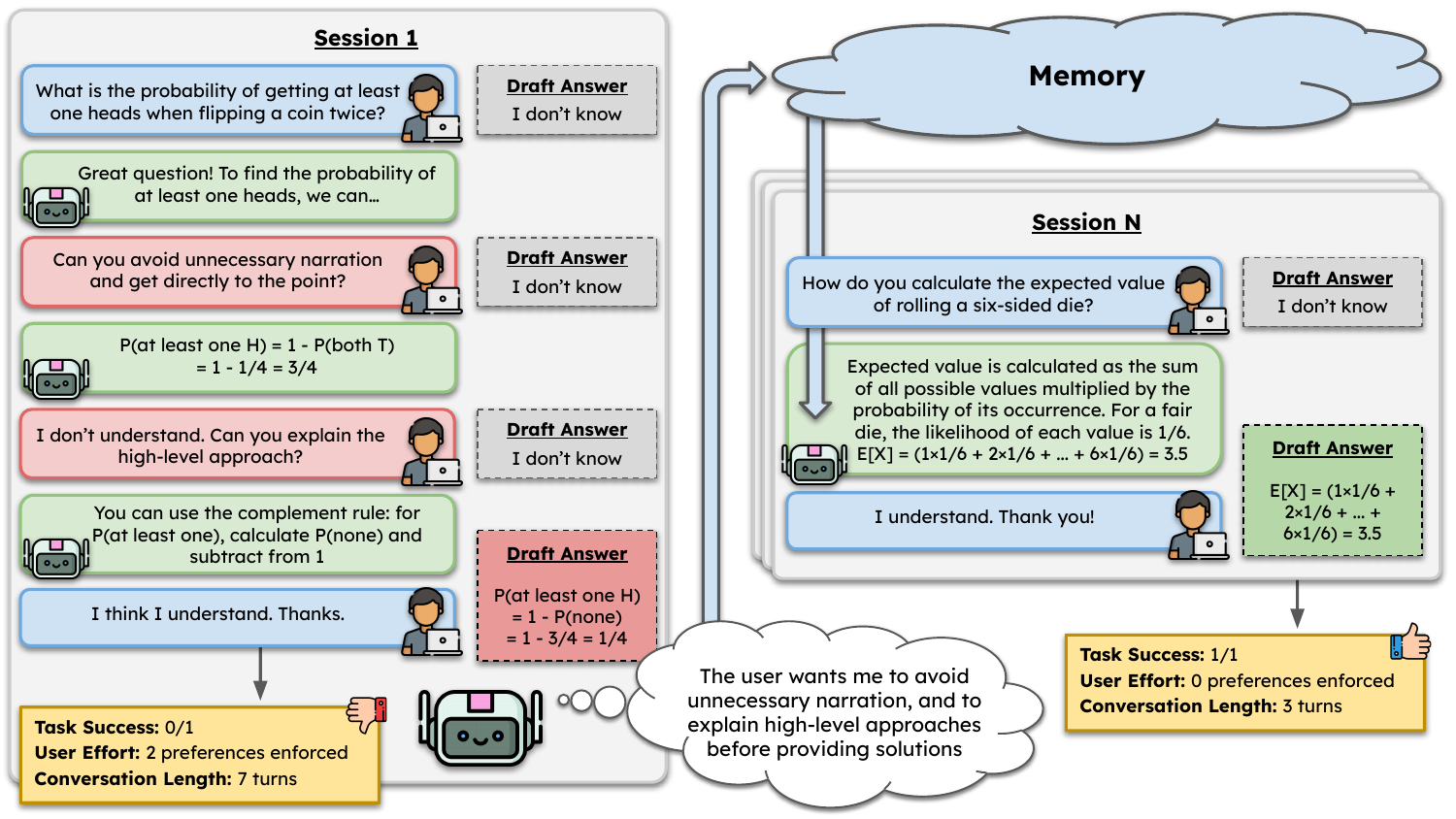}
    \vspace{-0.5em}
    \caption{The \textsc{MultiSessionCollab} benchmark with our long-term collaborative agent. Each session involves a user seeking help for a problem. The user maintains a draft answer that represents what they have learned from the interaction. They update their draft answer when the agent's responses are both helpful and preference-aligned. When responses violate preferences, the user enforces them, as indicated by the red text boxes. After each session, the agent reflects on the interaction to identify user preference information that will be useful for future interactions and update their memory accordingly. We measure collaboration quality in each session with task success, conversation length, and user effort.}
    \label{fig:multisessioncollab}
    \vspace{-1em}
\end{figure*}

Recent works have introduced memory mechanisms to overcome the finite context window of Large Language Models (LLMs) and enable agents to dynamically store and retrieve user-specific information \citep{chhikara2025mem0, wang2025mirix, yan2025memory}. Evaluations of these systems generally focus on how accurately agents can recall information and answer questions about past interactions \citep{xu-etal-2022-beyond, maharana2024evaluating, wu2025longmemeval}, rather than their downstream impact on long-term interactions with users. While recall is a necessary prerequisite, it does not capture whether agents can recognize what information is useful and leverage it to meaningfully improve future interactions.

To address this gap, we introduce \textsc{MultiSessionCollab}, a benchmark designed to evaluate how well conversational agents can learn user preference information and leverage it to meaningfully improve interactions during multi-session collaboration. It features user simulators that collaborate with the agent on problem-solving tasks across multiple sessions. Each user has distinct personas and interaction preferences that are derived from psychology, cognitive science, and human-computer interaction research, reflecting realistic human behaviors in collaborative settings (see taxonomy in Appendix \ref{appendix:interaction_preferences}) \citep{wood2007new, sweller2006worked, schwartz2002maximizing}. The benchmark is designed so that users progress most effectively when agent responses align with their preferences, incentivizing agents to continuously learn and adhere to preferences. We present an overview in in Figure \ref{fig:multisessioncollab}.

Unlike prior memory systems that optimize for information recall, we develop agents that are equipped with a more user-centric memory that learn user preferences to improve collaboration quality over time. As shown in Figure \ref{fig:multisessioncollab}, agents reflect on interactions to identify user preference information that will be valuable for future interactions and update their memory accordingly. During subsequent sessions, the full memory is provided to the agent, and relevant parts are dynamically retrieved throughout the interaction to guide agent behavior towards being more preference-aligned.

We also present a reinforcement learning (RL) framework that trains agents with learning signals from user simulator behavior in \textsc{MultiSessionCollab} to update memory more effectively. We use Group Relative Policy Optimization (GRPO) \citep{shao2024deepseekmath} with an LLM-judge that rewards reflections for comprehensively capturing user preferences revealed during interaction.

Extensive experiments across several models and five problem-solving tasks demonstrate that our memory architecture enables agents to continuously learn user interaction preferences and improve collaboration quality, yielding higher task success rates, more efficient interactions, and reduced user effort. Analyses of performance across sessions reveal continued improvement throughout the sessions, with the steepest gains occurring in early sessions before gradually stabilizing. Moreover, we show that our memory outperforms existing existing memory architectures at improving collaboration across sessions, and is even competitive with an oracle given direct access to ground-truth user preferences, highlighting the value of designing memory to be user-centric and learning through user interactions.

Finally, we conduct a human user study with 19 participants who engage in three consecutive collaborative sessions with an agent across coding, writing, and problem-solving tasks. Results align with our experimental findings and demonstrate that equipping agents with memory improves collaboration quality across sessions. Participants described these agents as more personalized and proactive, while also identifying challenges in cross-domain preference generalization.

\noindent The contributions of our work are:
\begin{itemize}[leftmargin=12pt,topsep=0pt]
\itemsep -0.5ex
\item We introduce \textsc{MultiSessionCollab} to evaluate how well conversational agents can learn user preferences and leverage them to improve interactions during multi-session collaboration.
\item We develop long-term collaborative agents equipped with memory that enables them to learn and leverage user preferences over time.
\item We present an RL framework that trains agents to generate more comprehensive reflections and update memory more effectively by using rewards derived from user behavior signals.
\item Through extensive experiments and a human user study, we demonstrate that our memory improves multi-session collaboration.
\end{itemize}
\section{Related Work}

\paragraph{Multi-Session Conversation Evaluation.} Early works that evaluate LLMs in multi-session interactions focus on how well models can generate responses that are consistent with past interactions \citep{xu-etal-2022-beyond, xu-etal-2022-long} and accurately update memory over time \citep{bae2022keep}. More recently, benchmarks have shifted towards question-answering style evaluations that assess how well LLMs can remember information from past interactions \citep{packer2023memgpt, maharana2024evaluating, wu2025longmemeval, hu2025evaluating, jiang2025know, zhao2025do}. In contrast, our work evaluates the downstream impact of memory on collaborative problem solving, incorporating abilities such as identifying useful information and actually leveraging them to improve user interactions.

\paragraph{Memory.} Providing LLMs with large contexts is computationally inefficient and can degrade performance, since LLMs struggle to effectively handle large amounts of information \citep{shi2023large, liu2024lost}. While one line of works try to improve LLM abilities to handle larger contexts \citep{liu2025comprehensive}, another line of works introduces memory to enable agents to store information from past experiences, and retrieve it when useful in future interactions \citep{shinn2023reflexion, packer2023memgpt, zhong2024memorybank, suzgun2025dynamic, wang2025agent, ho2025arcmemo, chhikara2025mem0}. More recent works introduce more sophisticated memory architectures, including those that leverage multi-agent systems \citep{wang2025mirix} or temporal-aware knowledge graphs \citep{rasmussen2025zep}. While these works demonstrate focus on memory for question-answering tasks, we present memory for improving user interactions during multi-session collaboration.

\paragraph{RL for Memory.} RL has been used to train agents to manage and utilize memory more effectively. Existing approaches use rewards based on question-answering correctness \citep{zhou2025mem1, yan2025memory, wang2025mem, yu2025memagent}. In contrast, we derive rewards from user behavior in interactions. Specifically, we train agents to recognize what user preference information revealed during an interaction will be valuable for improving future sessions.

\paragraph{Human-AI Collaboration.} Recent work has demonstrated that asking clarifying questions can enhance multi-turn interactions by helping agents better understand tasks, user intent, and preferences \citep{zhang2025modeling, li2025alfa, li2025eliciting, andukuri2024stargate, wan2025enhancing, li2025personalized}, with several works showing that this behavior is helpful for human-AI collaboration \citep{wu2025collabllm, zhou2025sweet, wang2025beyond}. These approaches focus on single-session interactions, which is appropriate for cold-start scenarios. However, as users increasingly engage with agents over multiple sessions, repeatedly asking the similar questions and eliciting preferences can become tedious and place unnecessary cognitive burden on users. Our work addresses this limitation by enabling agents remember user preference information across multiple sessions.
\section{\textsc{MultiSessionCollab}}
\label{sec:MultiSessionCollab}

We introduce \textsc{MultiSessionCollab} for evaluating conversational agents' ability to learn user preference information and leverage it to meaningfully improve collaboration quality over time. The benchmark takes place in a multi-session collaborative problem-solving setting with diverse user simulators, and can be instantiated with any problem-solving task. An overview of \textsc{MultiSessionCollab} is presented in Figure \ref{fig:multisessioncollab}.

\subsection{Collaborative Problem-Solving Session}
\label{sec:collaborative_problem_solving_session}
In each session, a user simulator seeks to solve a problem with assistance from a conversational agent. Following \citet{wu2025collabllm, zhou2025sweet}, only the user has access to the problem statement, and they start the conversation by providing an initial description of their problem. Over multiple turns, the agent asks clarifying questions to better understand the user's task and provides explanations to help the user develop a solution. Conversations last until the user is satisfied and decides to terminate, or when the maximum conversation length of 10 user-agent turns has been reached. This design mirrors realistic collaborative scenarios, such as tutoring or debugging, and results in natural multi-turn interactions.

In such collaborative settings, agents must not only provide correct information, but also communicate it in ways that users can understand, apply, and learn from \citep{shi2025models}. We capture this by having users maintain a \textit{draft answer} for their solution to the problem. The draft answer is never visible to the agent. It is initially empty, and is progressively updated as users receive assistance.

Crucially, users update their draft answer only when the agent provides helpful information in a manner that aligns with their interaction preferences. Information delivered in a way that violates the user's preferences (e.g. using overly technical language when the user prefers simple explanations) hinders their ability to effectively understand and apply the information \citep{shi2025models, sweller1988cognitive}. In such cases, users explicitly communicate their preferences to the agent rather than updating their draft answer. We note that real-world users may be able to extract useful information even when responses do not fully align with their preferences. Our design choice of not updating the draft answer in such cases helps isolate preference adherence, enabling more direct evaluation of how well agents can learn and adapt to preferences across sessions.

\subsection{User Simulator Design}
\label{sec:UserSimulatorDesign}
A central component of our benchmark is designing user simulators with realistic user profiles that exhibit diverse behaviors and interaction preferences.

First, we instantiate each user profile with a randomly selected persona from Persona Hub, a large-scale persona collection \citep{ge2024scaling}. Each persona exhibits unique knowledge, experiences, interests, personalities and professions, which translates to varying perspectives and diverse behaviors.

Each user profile is also assigned a random set of three interaction preferences that describe how they expect the agent to behave during collaboration. The preferences may specify specific communication styles \citep{miehle2020estimating}, learning approaches \citep{chi1989self}, or proactivity levels \citep{horvitz1999principles}. For instance, a user may prefer concise responses with key takeaways highlighted at the end, or expect detailed step-by-step explanations. Each preference is grounded in studies from psychology, cognitive science, and human-computer interaction, and reflects realistic human behavior in problem-solving settings. A complete taxonomy of the preferences and their sources are provided in Appendix \ref{appendix:interaction_preferences}.

\subsection{Evaluation Metrics}
\label{multisessioncollab:metrics}
For each session, we evaluate collaboration quality across three dimensions:
\begin{itemize}[leftmargin=12pt,topsep=0pt]
\itemsep -0.5ex
    \item \textit{Task Success}: the accuracy of the user's final draft answer, which represents what the user learned from the interaction. Task success is a standard metric for collaboration \citep{wu2025collabllm, wang2025beyond}.
    \item \textit{User Effort}: the number of times the user enforced their preferences. Each enforcement represents a preference adherence failure, placing cognitive burden on the user to correct the agent behavior rather than focus on the task.
    \item \textit{Conversation Length}: the total number of turns in the conversation. Agents that adhere to preferences reduce friction and help users complete the task in fewer turns.
\end{itemize}
Higher task success, lower user effort, and shorter conversations indicate more effective collaboration.

\subsection{Multi-Session Setting} 
For each user, we sequentially run multiple collaborative problem-solving sessions, where each session involves a different problem. The agent initially has no knowledge of user preferences, and is expected to learn and adapt to preferences throughout the sessions. We report metrics averaged across all sessions per user, and then across all users.
\section{Methodology}
\subsection{Long-Term Collaborative Agents}

Conversational agents start with no knowledge about user preferences and inevitably generate misaligned responses during early interactions. But as interaction experience accumulates, users reveal their preferences, and long-term collaborative agents must be able to learn from these signals to reduce cognitive burden on users and improve collaboration quality over time.

To build such agents, we equip them with a memory architecture that is specifically designed to persist and refine user preference information across sessions and improve interactions. It is illustrated in Figure \ref{fig:multisessioncollab} and all prompts are provided in Appendix \ref{appendix:agent_prompts}. The memory architecture comprises two components:
\begin{enumerate}[leftmargin=12pt,topsep=0pt]
\itemsep -0.5ex
    \item \textbf{Session-Level Reflection:} the agent analyzes the interaction and identifies preference information that would be useful for improving future interactions, which can include the specific preferences themselves, the contexts in which they apply, and details about which actions or approaches satisfy each preference. The agent then updates their existing memory with this new information.
    \item \textbf{Persistent Memory:} the memory is provided at the start of each session as part of the agent's system prompt, enabling them to adapt their behavior according to learned preferences without requiring users to repeatedly specify them. Additionally, at each turn, we use an LLM to analyze the conversation and retrieve specific parts of the memory that are directly relevant to the conversational context and provide them to the agent.
\end{enumerate}

\subsection{Reinforcement Learning in \textsc{MultiSessionCollab}}
\label{methodology:rl}

\begin{figure}[]
    \centering
    \includegraphics[trim=0 0 0pt 0pt, clip, width=0.9\linewidth]{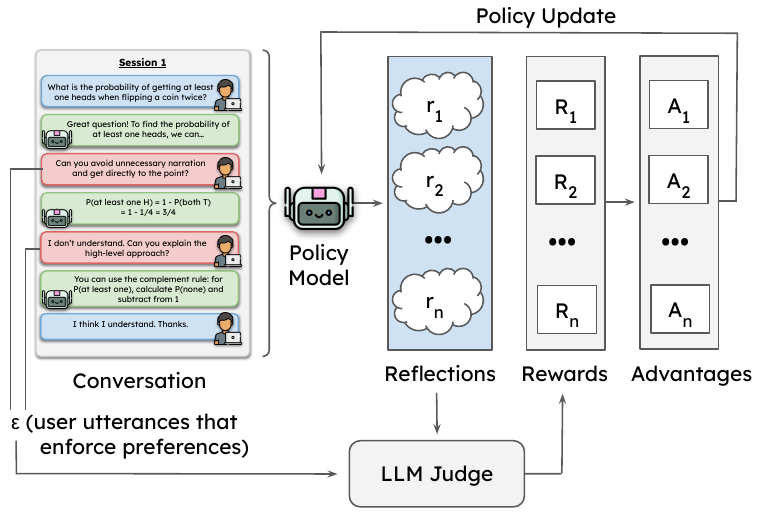}
    \caption{RL framework for improving session-level reflections. The policy model generates $n$ reflection rollouts for a conversation. The judge model evaluates each reflection against $\varepsilon$ (the subset of user utterances that enforce preferences) and assigns rewards. Advantages are computed and the policy is updated via GRPO.}
    \label{fig:grpo}
    \vspace{-1em}
\end{figure}

We present an RL framework that uses the \textsc{MultiSessionCollab} environment to train agents to generate more comprehensive session-level reflections that effectively capture the user preferences revealed during interactions (Figure \ref{fig:grpo}). More specifically, we use learning signals derived from user simulator behavior to design rewards for the GRPO algorithm.

\paragraph{Problem Formulation. } For a collaborative session, let $C=\{ (u_1, a_1), ..., (u_n, a_n) \}$ denote a conversation consisting of $n$ turns, where $(u_i, a_i)$ are the user and agent utterances at turn $i$. Let $\varepsilon = \{ e_1, ..., e_k \} \subset \{ u_1, ..., u_n \}$ denote the subset of utterances where the user enforces their preferences. Given a conversation $C$, the agent must generate a reflection $r$ that captures all preference information revealed during the interaction.

\paragraph{Reward Design. } A session-level reflection is considered successful if it comprehensively captures the user preferences revealed during the interaction and provides actionable guidance for future sessions. $\varepsilon$ specifies all information that a reflection should capture. We therefore use an LLM judge to evaluate how well $r$ covers all preferences in $\varepsilon$ without any hallucinations, and get an objective scalar reward $R_{coverage}(r,\varepsilon)$. We also include a format reward $R_{format}(r)$ that encourages well-structured outputs containing both a reasoning trace and the reflection. The total reward is:
\begin{equation}
    R(r,\varepsilon) = R_{coverage}(r,\varepsilon) + R_{format}(r)
\end{equation}

The LLM judge prompts are provided in Appendix \ref{appendix:agent_prompts}. Using this reward, we employ GRPO to train a policy $\pi_{\theta}$ to generate more comprehensive session-level reflections that have no hallucinations. As a result, the agent is able to make more meaningful memory updates that helps them be more preference-aligned during future interactions.
\section{Experimental Setup}
\label{sec:experimental_setup}

To rigorously evaluate conversational agents with \textsc{MultiSessionCollab}, we instantiate the benchmark with 100 user profiles and five problem-solving benchmarks that cover diverse domains: MATH-500 \citep{hendrycks2021measuringmath}, MATH-Hard \citep{hendrycks2021measuringmath}, LogiQA \citep{liu2007logiqa}, MMLU \citep{hendrycks2021measuring}, and MedQA \citep{jin2021disease}. We use Llama-3.3-70B-Instruct \citep{dubey2024llama} for our user simulator and LLM-judge. Each user collaborates with the agent to solve 20 randomly sampled problems, with one problem per session and a maximum of 10 conversational turns per session. Across all benchmarks, this totals 10,000 collaborative sessions per agent, ensuring robust evaluation of agent performance during multi-session collaboration.

\paragraph{Environment Validation.} To validate the reliability of \textsc{MultiSessionCollab} and understand the contribution of different components, we conduct a series of ablation studies that isolate the impact of various dimensions. We observe a drop in performance when moving from direct problem-solving to multi-turn interaction, which aligns with prior work showing that LLMs struggle to ask clarifying questions or maintain coherence across turns \citep{wu2025collabllm, laban2025llms, zhou2025sweet, mehri2025goal}. Performance decreases further when introducing user preferences, since users correct agent behavior during preference violations rather than focus on the task. Our complete analysis is provided in Appendix \ref{appendix:env_validation}. We also present a user study that confirms our findings generalize to real users in Section \ref{sec:user_study}.

\paragraph{Training.} Training data is constructed using a different set of user profiles and problems, with no overlap with evaluation data. Similar to evaluation, we have 100 user profiles and select 20 problems from each problem-solving benchmark. Using Llama-3.3-70B-Instruct as both the user simulator and the agent, we generate 10,000 collaborative sessions along with reflections. 

We train Llama-3.1-8B-Instruct \citep{dubey2024llama} and Qwen-2.5-7B-Instruct \citep{qwen2025qwen25technicalreport} to generate session-level reflections given a conversation. The models are first initialized with supervised fine-tuning (SFT) as a cold start, which helps stabilizes RL optimization and leads to more effective performance \citep{guo2025deepseek}. Then, we conduct GRPO as described in Section \ref{methodology:rl}. Details on hyperparameters and issues such as catastrophic forgetting are provided in Appendix \ref{appendix:training}.

We evaluate several LLMs in \textsc{MultiSessionCollab}, including Llama-3.1-8B-Instruct, Qwen-2.5-7B-Instruct, gpt-oss-20b, and Llama-3.3-70B-Instruct \citep{qwen2025qwen25technicalreport, openai2025gptoss120bgptoss20bmodel, dubey2024llama}. Models are evaluated as standard conversational agents without memory, and also when equipped with memory. For Llama-3.1-8B-Instruct and Qwen-2.5-7B-Instruct models, we additionally report results after applying GRPO training to improve session-level reflections.
\section{Results}
\label{sec:results}

\definecolor{darkgreen}{rgb}{0.0, 0.5, 0.0}
\definecolor{darkred}{rgb}{0.5, 0.0, 0.0}

\definecolor{mygrey}{gray}{0.90} 
\definecolor{mygrey2}{gray}{0.80}
\definecolor{lightblue}{RGB}{220, 235, 250}

\begin{table*}[t]
    \centering
    \scriptsize
    \setlength{\tabcolsep}{4pt}

    \begin{tabular}{c l | ccc | ccc | ccc} 
        \toprule
        \multicolumn{2}{c}{\multirow{2}[3]{*}{}}
        & \multicolumn{3}{c}{\textbf{MATH-500}} & \multicolumn{3}{c}{\textbf{MATH-Hard}} & \multicolumn{3}{c}{\textbf{LogiQA}} \\ 

        \cmidrule(lr){3-5} \cmidrule(lr){6-8} \cmidrule(lr){9-11}

        & & \textit{TS (\%)} $\uparrow$ & \textit{UE} $\downarrow$ & \textit{Len} $\downarrow$ & \textit{TS (\%)} $\uparrow$ & \textit{UE} $\downarrow$ & \textit{Len} $\downarrow$ & \textit{TS (\%)} $\uparrow$ & \textit{UE} $\downarrow$ & \textit{Len} $\downarrow$ \\ 

        \midrule

        & Qwen-2.5-7B    & 51.35 & 2.95 & 14.05     & 26.5 & 2.93 &  14.62  & 19.65 & 2.40 &  15.61 \\

        \rowcolor{mygrey} & \quad + memory      & 51.75$_{\textcolor{darkgreen}{\uparrow 0.40}}$ & 2.80$_{\textcolor{darkgreen}{\downarrow 0.15}}$ & 13.46$_{\textcolor{darkgreen}{\downarrow 0.59}}$     & 26.21$_{\textcolor{darkred}{\downarrow 0.29}}$ & 2.86$_{\textcolor{darkgreen}{\downarrow 0.07}}$ & 14.40$_{\textcolor{darkgreen}{\downarrow 0.22}}$         & 18.61$_{\textcolor{darkred}{\downarrow 1.04}}$ & 2.35$_{\textcolor{darkgreen}{\downarrow 0.05}}$  & 15.57$_{\textcolor{darkgreen}{\downarrow 0.04}}$ \\

        \rowcolor{mygrey2} & \quad \quad + GRPO      & 56.80$_{\textcolor{darkgreen}{\uparrow 5.45}}$ & 2.16$_{\textcolor{darkgreen}{\downarrow 0.79}}$ & 12.85$_{\textcolor{darkgreen}{\downarrow 1.20}}$     & 27.75$_{\textcolor{darkgreen}{\uparrow 1.25}}$ & 2.27$_{\textcolor{darkgreen}{\downarrow 0.66}}$ & 13.78$_{\textcolor{darkgreen}{\downarrow 0.84}}$         & 23.20$_{\textcolor{darkgreen}{\uparrow 3.55}}$ & 1.74$_{\textcolor{darkgreen}{\downarrow 0.66}}$  & 15.01$_{\textcolor{darkgreen}{\downarrow 0.60}}$ \\

        \midrule

        & Llama-3.1-8B    & 44.55 & 2.73 & 14.83     & 11.50 & 2.84 &  16.15  & 24.45 & 2.44 &  16.50 \\

        \rowcolor{mygrey} & \quad + memory      & 44.93$_{\textcolor{darkgreen}{\uparrow 0.38}}$ & 2.20$_{\textcolor{darkgreen}{\downarrow 0.53}}$ & 13.73$_{\textcolor{darkgreen}{\downarrow 1.10}}$     & 12.79$_{\textcolor{darkgreen}{\uparrow 1.29}}$ & 2.33$_{\textcolor{darkgreen}{\downarrow 0.51}}$ & 15.28$_{\textcolor{darkgreen}{\downarrow 0.87}}$         & 26.50$_{\textcolor{darkgreen}{\uparrow 2.05}}$ & 1.92$_{\textcolor{darkgreen}{\downarrow 0.52}}$  & 15.93$_{\textcolor{darkgreen}{\downarrow 0.57}}$ \\

        \rowcolor{mygrey2} & \quad \quad + GRPO      & 48.68$_{\textcolor{darkgreen}{\uparrow 4.13}}$ & 1.85$_{\textcolor{darkgreen}{\downarrow 0.88}}$ & 12.85$_{\textcolor{darkgreen}{\downarrow 1.98}}$     & 13.59$_{\textcolor{darkgreen}{\uparrow 2.09}}$ & 2.01$_{\textcolor{darkgreen}{\downarrow 0.83}}$ & 14.57$_{\textcolor{darkgreen}{\downarrow 1.58}}$         & 26.55$_{\textcolor{darkgreen}{\uparrow 2.10}}$ & 1.63$_{\textcolor{darkgreen}{\downarrow 0.81}}$  & 15.57$_{\textcolor{darkgreen}{\downarrow 0.93}}$ \\

        \midrule

        & gpt-oss-20b    & 67.60 & 2.90 & 13.95    & 41.45 & 3.46 & 15.73  & 20.31 & 2.90 & 16.56 \\

        \rowcolor{mygrey2} & \quad + memory      & 70.35$_{\textcolor{darkgreen}{\uparrow 2.75}}$ & 2.34$_{\textcolor{darkgreen}{\downarrow 0.56}}$ & 12.72$_{\textcolor{darkgreen}{\downarrow 1.23}}$     & 45.07$_{\textcolor{darkgreen}{\uparrow 3.62}}$ & 2.83$_{\textcolor{darkgreen}{\downarrow 0.63}}$ & 14.33$_{\textcolor{darkgreen}{\downarrow 1.40}}$   & 25.19$_{\textcolor{darkgreen}{\uparrow 4.88}}$ & 2.32$_{\textcolor{darkgreen}{\downarrow 0.58}}$ & 14.89$_{\textcolor{darkgreen}{\downarrow 1.67}}$ \\

        \midrule

        & Llama-3.3-70B    & 59.29 & 3.00 & 14.98    & 25.81 & 3.27  & 17.03  & 26.69 & 2.96  & 17.08 \\

        \rowcolor{mygrey2} & \quad + memory      & 63.85$_{\textcolor{darkgreen}{\uparrow 4.56}}$ & 1.99$_{\textcolor{darkgreen}{\downarrow 1.01}}$ & 13.28$_{\textcolor{darkgreen}{\downarrow 1.70}}$     & 30.50$_{\textcolor{darkgreen}{\uparrow 4.69}}$ & 2.25$_{\textcolor{darkgreen}{\downarrow 1.02}}$ & 15.17$_{\textcolor{darkgreen}{\downarrow 1.86}}$   & 30.25$_{\textcolor{darkgreen}{\uparrow 3.56}}$ & 1.89$_{\textcolor{darkgreen}{\downarrow 1.07}}$ & 15.85$_{\textcolor{darkgreen}{\downarrow 1.23}}$ \\

    \end{tabular}

    \vspace{1em}

    \begin{tabular}{c l | ccc | ccc || ccc ||}
        \toprule
        \multicolumn{2}{c}{\multirow{2}[3]{*}{}}
        & \multicolumn{3}{c}{\textbf{MMLU}} & \multicolumn{3}{c}{\textbf{MedQA}} & \multicolumn{3}{c}{\fcolorbox{black}{lightblue}{\textbf{Overall}}} \\

        \cmidrule(lr){3-5} \cmidrule(lr){6-8} \cmidrule(lr){9-11}

        & & \textit{TS (\%)} $\uparrow$ & \textit{UE} $\downarrow$ & \textit{Len} $\downarrow$ & \textit{TS (\%)} $\uparrow$ & \textit{UE} $\downarrow$ & \textit{Len} $\downarrow$ & \textit{TS (\%)} $\uparrow$ & \textit{UE} $\downarrow$ & \textit{Len} $\downarrow$ \\ 

        \midrule

        & Qwen-2.5-7B     & 47.85 & 2.51  & 13.81     & 35.70 & 2.58  & 15.96  & 36.21 & 2.67 & 14.81 \\
        
        \rowcolor{mygrey} & \quad + memory      & 45.17$_{\textcolor{darkred}{\downarrow 2.68}}$  & 2.46$_{\textcolor{darkgreen}{\downarrow 0.05}}$ & 13.72$_{\textcolor{darkgreen}{\downarrow 0.09}}$     & 34.16$_{\textcolor{darkred}{\downarrow 1.54}}$ & 2.65$_{\textcolor{darkred}{\uparrow 0.07}}$ & 16.33$_{\textcolor{darkred}{\uparrow 0.37}}$ & 35.18$_{\textcolor{darkred}{\downarrow 1.03}}$ & 2.62$_{\textcolor{darkgreen}{\downarrow 0.05}}$ & 14.70$_{\textcolor{darkgreen}{\downarrow 0.11}}$ \\

        \rowcolor{mygrey2} & \quad \quad + GRPO      & 51.15$_{\textcolor{darkgreen}{\uparrow 3.30}}$  & 1.91$_{\textcolor{darkgreen}{\downarrow 0.60}}$ & 12.88$_{\textcolor{darkgreen}{\downarrow 0.93}}$     & 39.30$_{\textcolor{darkgreen}{\uparrow 3.60}}$ & 1.97$_{\textcolor{darkgreen}{\downarrow 0.61}}$ & 15.61$_{\textcolor{darkgreen}{\downarrow 0.35}}$ & 39.64$_{\textcolor{darkgreen}{\uparrow 3.43}}$ & 2.01$_{\textcolor{darkgreen}{\downarrow 0.66}}$ & 14.02$_{\textcolor{darkgreen}{\downarrow 0.79}}$ \\

        \midrule

        & Llama-3.1-8B     & 50.50 & 2.42  & 14.12     & 39.20 & 2.74  & 16.66     & 34.04 & 2.63  & 15.65 \\
        
        \rowcolor{mygrey} & \quad + memory      & 51.50$_{\textcolor{darkgreen}{\uparrow 1.00}}$  & 2.00$_{\textcolor{darkgreen}{\downarrow 0.42}}$ & 13.62$_{\textcolor{darkgreen}{\downarrow 0.50}}$     & 41.85$_{\textcolor{darkgreen}{\uparrow 2.65}}$ & 2.10$_{\textcolor{darkgreen}{\downarrow 0.64}}$ & 16.00$_{\textcolor{darkgreen}{\downarrow 0.66}}$     & 35.51$_{\textcolor{darkgreen}{\uparrow 1.47}}$ & 2.11$_{\textcolor{darkgreen}{\downarrow 0.52}}$ & 14.91$_{\textcolor{darkgreen}{\downarrow 0.74}}$ \\

        \rowcolor{mygrey2} & \quad \quad + GRPO      & 54.00$_{\textcolor{darkgreen}{\uparrow 3.50}}$  & 1.68$_{\textcolor{darkgreen}{\downarrow 0.42}}$ & 12.92$_{\textcolor{darkgreen}{\downarrow 1.20}}$     & 40.95$_{\textcolor{darkgreen}{\uparrow 1.75}}$ & 2.10$_{\textcolor{darkgreen}{\downarrow 0.64}}$ & 15.89$_{\textcolor{darkgreen}{\downarrow 0.77}}$     & 36.75$_{\textcolor{darkgreen}{\uparrow 2.71}}$ & 1.85$_{\textcolor{darkgreen}{\downarrow 0.78}}$ & 14.36$_{\textcolor{darkgreen}{\downarrow 0.45}}$ \\

        \midrule

        & gpt-oss-20b    & 49.62 & 2.46 & 13.81    & 32.75 & 3.26 & 17.06  & 42.35 & 3.00 & 15.42 \\

        \rowcolor{mygrey2} & \quad + memory      & 55.05$_{\textcolor{darkgreen}{\uparrow 5.43}}$ & 2.0$_{\textcolor{darkgreen}{\downarrow 0.46}}$ & 12.23$_{\textcolor{darkgreen}{\downarrow 1.58}}$     & 31.68$_{\textcolor{darkred}{\downarrow 1.07}}$ & 3.76$_{\textcolor{darkred}{\uparrow 0.50}}$ & 16.70$_{\textcolor{darkgreen}{\downarrow 0.36}}$   & 45.47$_{\textcolor{darkgreen}{\uparrow 3.12}}$ & 2.65$_{\textcolor{darkgreen}{\downarrow 0.35}}$ & 14.17$_{\textcolor{darkgreen}{\downarrow 1.25}}$ \\

        \midrule

        & Llama-3.3-70B    & 51.26 & 2.76 & 14.34    & 45.84 & 2.91 & 16.38  & 41.78 & 2.98 & 15.96\\

        \rowcolor{mygrey2} & \quad + memory      & 57.35$_{\textcolor{darkgreen}{\uparrow 6.09}}$ & 1.74$_{\textcolor{darkgreen}{\downarrow 1.02}}$ & 12.99$_{\textcolor{darkgreen}{\downarrow 1.35}}$     & 49.95$_{\textcolor{darkgreen}{\uparrow 4.11}}$ & 1.92$_{\textcolor{darkgreen}{\downarrow 0.99}}$ & 15.35$_{\textcolor{darkgreen}{\downarrow 1.03}}$     & 46.38$_{\textcolor{darkgreen}{\uparrow 4.60}}$ & 1.96$_{\textcolor{darkgreen}{\downarrow 1.02}}$ & 14.53$_{\textcolor{darkgreen}{\downarrow 1.43}}$ \\

        \bottomrule
    \end{tabular}

    \caption{Conversational agent performance on \textsc{MultiSessionCollab} across five problem-solving tasks. We report task success (\textit{TS}), user effort (\textit{UE}), and conversation length (\textit{Len}). For each model, the first row is the baseline conversational agent without memory, \textit{+memory} is the agent equipped with our memory architecture, and \textit{+GRPO} is the agent trained to generate session-level reflections. Subscripts denote change relative to the baseline. Overall performance is also reporting by averaging performance across all tasks. We use the instruct variant for all models.}

    \label{tab:results}
    \vspace{-1em}
\end{table*}
\begin{table}[t]
    \centering
    \small
    \setlength{\tabcolsep}{5pt}
    \begin{tabular}{l | ccc}
        \toprule
        & \multicolumn{3}{c}{\fcolorbox{black}{lightblue}{\textbf{Overall}}} \\
        \cmidrule(lr){2-4}
        & \textit{TS (\%)} $\uparrow$ & \textit{UE} $\downarrow$ & \textit{Len} $\downarrow$ \\
        \midrule
        Llama-3.3-70B & 41.78 & 2.98 & 15.96 \\
        \rowcolor{mygrey} \quad + Mem0 & 43.27$_{\textcolor{darkgreen}{\uparrow 1.49}}$ & 2.19$_{\textcolor{darkgreen}{\downarrow 0.79}}$ & 15.06$_{\textcolor{darkgreen}{\downarrow 0.90}}$ \\
        \rowcolor{mygrey2} \quad + memory (ours) & 46.38$_{\textcolor{darkgreen}{\uparrow 4.60}}$ & 1.96$_{\textcolor{darkgreen}{\downarrow 1.02}}$ & 14.53$_{\textcolor{darkgreen}{\downarrow 1.43}}$ \\
        \bottomrule
    \end{tabular}
    \caption{Comparison between our memory architecture and Mem0 \citep{chhikara2025mem0} on \textsc{MultiSessionCollab}. We report task success (\textit{TS}), user effort (\textit{UE}), and conversation length (\textit{Len}) averaged across all tasks. Subscripts denote change over having no memory.}
    \label{tab:mem0_comparison}
    \vspace{-1.65em}
\end{table}

\paragraph{Main Results.}

Table \ref{tab:results} presents our experimental results on the \textsc{MultiSessionCollab} benchmark. Equipping conversational agents with our memory architecture enables them to learn preferences as they accumulate experience and proactively adhere to them without requiring users to repeatedly enforce them, reducing friction and allowing users to focus on the task. This results in improvements in collaboration quality: interactions become more efficient, require less user effort, and yield higher task success rates.

We observe even further improvements when using our RL framework to train agents to generate session-level reflections that effectively capture the preferences revealed during interactions. We train Qwen-2.5-7B-Instruct and Llama-3.1-8B-Instruct, which leads to both models being able to leverage memory more effectively and notable improvements across all metrics. In particular, while Qwen-2.5-7B-Instruct initially exhibited a $1.03\%$ decrease in task success when equipped with memory, after training, the same model achieves a $3.43\%$ improvement in task success. This demonstrates that our memory architecture's effectiveness depends largely on reflection quality. Overall, these results suggest that RL with learning signals derived from user simulator behavior offers a promising direction for designing memory agents for improving user interactions.

\paragraph{Comparison with Existing Memory. } Unlike existing memory architectures, which are primarily designed for information retrieval and question-answering, ours is designed to improve interactions with users. To demonstrate this, we compare against Mem0, a state-of-the-art memory architecture that stores information in a vector database after each user-agent turn and dynamically retrieves them prior to response generation \citep{chhikara2025mem0}. We equip Llama-3.3-70B-Instruct with each memory architecture and present results averaged across all tasks in Table \ref{tab:mem0_comparison}.

Our memory architecture outperforms Mem0 across all metrics. We attribute this to a fundamental difference in what each memory captures. Mem0's memories tend to store factual information relevant to answering questions, whereas our reflections capture user-centric information about how to interact with the user. For example, a Mem0 memory is: \textit{"Evaluating the effectiveness of communication with the parents is a crucial step…"} -- a factual detail extracted from a previous response focused on answering the user's question. In contrast, a memory from our architecture is: \textit{"The user prefers a narrative and engaging style, with multiple viable approaches to addressing the problem..."}. These results highlight that for agents to develop long-term relationships with users and improve interactions over time, memory architectures must be explicitly designed to be more user-centric.

We additionally verify in Appendix \ref{appendix:memory_vs_full_history} that providing the full multi-session conversation history as context is not a viable alternative to memory.

\begin{figure}[]
    \centering
    \includegraphics[width=0.7\linewidth]{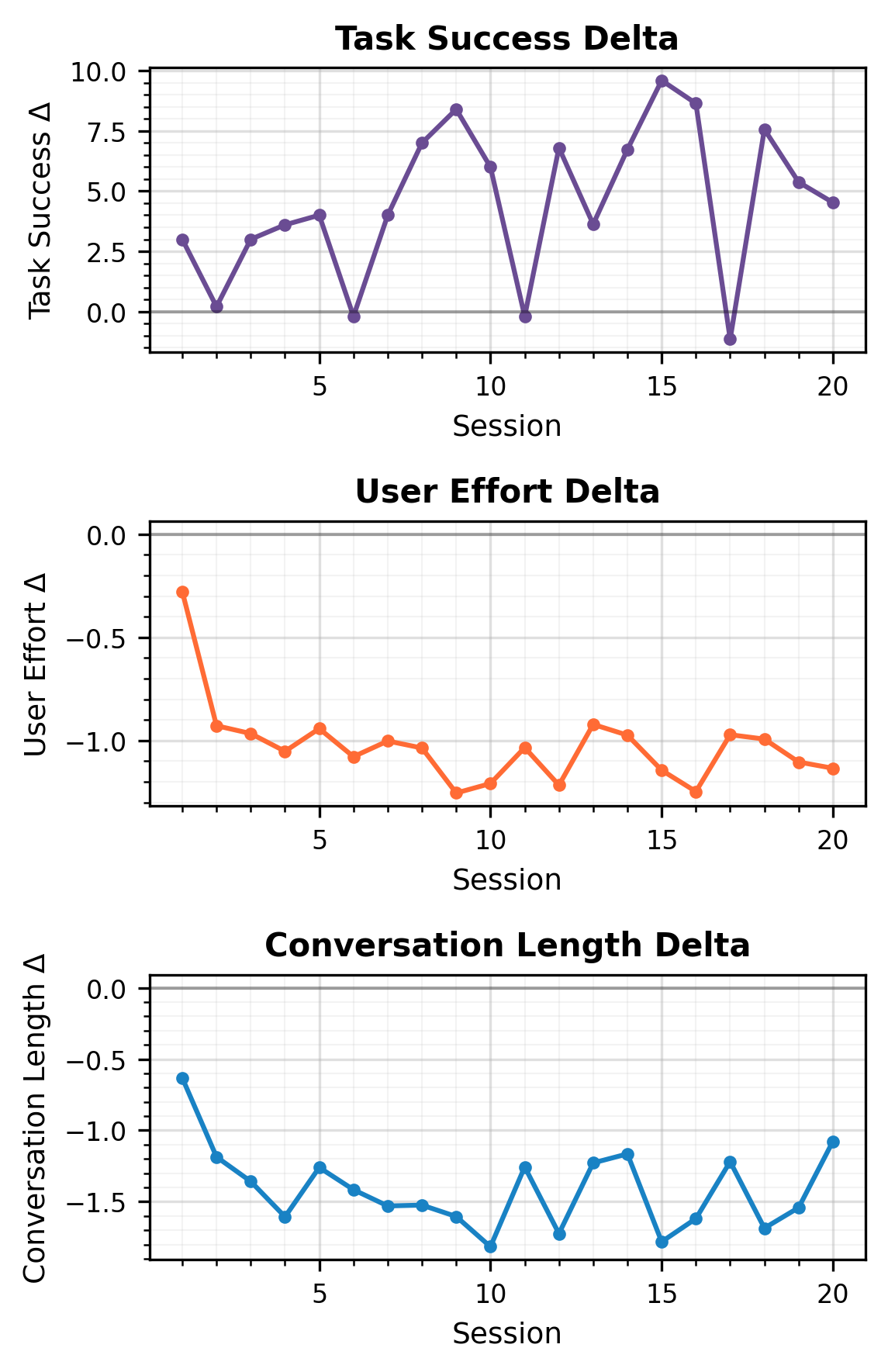}
    \caption{Performance across sessions for Llama-3.3-70B-Instruct. Each graph plots the delta between agents with memory and agents without memory across 20 sessions for Task Success ($\Delta_i^{TS}$) $\uparrow$, User Effort ($\Delta_i^{UE}$) $\downarrow$, and Conversation Length ($\Delta_i^{Len}$) $\downarrow$.}
    \label{fig:learning_dynamics_llama70b}
    \vspace{-1em}
\end{figure}

\paragraph{Performance Across Sessions.} Next, we analyze our agent's performance as they accumulate interaction experience throughout sessions. For each session $i \in \{1, ..., 20\}$, we compute the agent's average performance across all users and tasks for each metric. Then, we calculate the deltas $\Delta_i^{TS}$, $\Delta_i^{UE}$, and $\Delta_i^{Eff}$, where each delta denotes the difference between the agent with memory and the agent without memory at session $i$ for the respective metric. These deltas isolate the impact of memory at each session. Figure \ref{fig:learning_dynamics_llama70b} plots these deltas across sessions for Llama-3.3-70B-Instruct, with additional plots for other models presented in Appendix \ref{appendix:performance_across_sessions}. Increasing  $\Delta_i^{TS}$, and decreasing $\Delta_i^{UE}$ and $\Delta_i^{Eff}$ indicate that the agent is learning to collaboration more effectively.

We observe consistent trends of improvement across all deltas. $\Delta_i^{TS}$ has an upward trajectory, particularly after smoothing. However, there are also some points in the graph where the delta is close to 0 and the memory does not seem to help improve performance for these problems. $\Delta_i^{UE}$ and $\Delta_i^{Eff}$ show consistent downward trends, with the steepest improvements occurring within the first five sessions, and gradually stabilizing towards the last 10 sessions. This demonstrates how agents with memory \textit{continuously learn and refine their knowledge} as interaction experience accumulates, without plateauing in early sessions. 

Some variance in the results comes from certain preferences that make problem-solving more difficult, reflecting a limitation in the agent's ability to adhere to preferences during interactions while maintaining task performance.

\paragraph{Value of Learning Through Interaction.} We conduct an ablation study where agents are provided with direct access to user preferences, the same as what is given to the user simulator. This was originally intended to be an oracle to establish an upper bound on performance. Surprisingly, agents equipped with our memory architecture achieve competitive performance with this oracle, and sometimes outperform it, despite beginning with no information about the user. Our qualitative analysis indicates that our memory helps capture richer preference information from interactions, such as the contexts in which the preferences apply and specific strategies for satisfying them. Moreover, this information is continually refined across sessions as agents accumulate interaction experience. These results demonstrate the effectiveness of using memory to learn through actual user interaction. We report this and several additional ablations that help us better understand the memory architecture in Appendix \ref{appendix:env_validation}.
\section{Real-World User Study}
\label{sec:user_study}

\paragraph{Setup.} To complement our main experiments, we conduct a user study with 19 participants to understand the impact of memory in more realistic collaboration settings. The study lasted approximately 1.5 hours per participant, and was declared exempt by our Institutional Review Board (IRB). A comprehensive description of our study design and results are provided in Appendix \ref{appendix:user_study}.

We adopt a similar setup to \textsc{MultiSessionCollab} and design two study types: single-domain (coding only) and mixed-domain (writing, math problem-solving, and coding). In each study, participants complete three sequential sessions, solving one problem per session, with both a standard agent and an agent equipped with memory. Participants adopt a set of interaction preferences that they ensure the agents adhere to throughout the sessions. These preferences are parallel to those from Appendix \ref{appendix:interaction_preferences}. After each session, participants complete a survey where they rate the intrinsics aspects of the collaboration on a 5-point Likert scale: preference adherence, preference retention, confidence, and overall satisfaction.

\paragraph{Results.} Our user study results align with the trends observed in our simulated experiments. Across both single-domain and mixed-domain settings, agents with memory demonstrated consistent improvements over sessions: conversations required fewer turns, preference adherence and retention improved, and confidence and satisfaction increased (see Figure \ref{fig:user_study_graphs}). For instance, in the first session conversations with memory had a median of 8 (coding) and 6 (mixed) turns, and those without memory had a median of 8 turns. By the third session, agents with memory dropped to 6 and 4 turns, compared to 10 and 8 turns without memory.

Participants also provided free-form feedback after each session and shared overall thoughts at the end of the study. They described the agents with memory as noticeably more personalized, highlighting their ability to proactively adhere to preferences and handle vague queries. However, participants also noted limitations: personalization felt less effective in mixed-domain settings, and the agent's learned preferences were not as effective as when explicitly restated within a session. These findings suggest that while memory substantially improves multi-session collaboration, challenges remain in cross-domain generalization and fully capturing user preferences from interaction.
\section{Conclusion}
We introduce \textsc{MultiSessionCollab}, a benchmark for evaluating conversational agents' ability to learn user preferences and leverage them to meaningfully improve collaboration across multiple sessions. We develop long-term collaborative agents equipped with memory designed to learn preferences to improve interaction quality over time. We further demonstrate that \textsc{MultiSessionCollab} can serve as an effective RL environment, where we use learning signals derived from user behavior to encourage more comprehensive reflections, leading to more meaningful memory updates. Extensive experiments and user studies show that equipping agents with our memory improves collaboration quality across sessions, yielding higher task success rates, more efficient interactions, and reduced user effort.

\section{Limitations}
Our work provides a foundation for multi-session collaboration, with many promising directions for future research. We use Llama-3.3-70B-Instruct as our user simulator. While it is a highly capable model, user simulation requires reliably following complex instructions, such as consistently enforcing preferences throughout interactions, and our evaluation metrics depend on this behavior. We found that the simulator struggled to exhibit expected behaviors during more complex tasks such as coding, which we consequently excluded from our benchmark. Additionally, user simulators can be extended to capture more realistic human behaviors. For example, real users can express preferences implicitly through behavioral cues rather than explicitly. Or users can have more complex preferences, such as context-dependent preferences that vary across domains, or evolving preferences that change over time. Although we conduct a human user study to validate our findings across realistic tasks and real users, we acknowledge these limitations remain.

\bibliography{custom}

\clearpage

\appendix

\section{User Interaction Preferences}
\label{appendix:interaction_preferences}

In human-AI collaborative settings, users often have specific interaction preferences that describe how they expect agents to behave. When agents adhere to these preferences, users may learn more effectively, communicate more efficiently, or experience smoother, more personalized interactions \citep{shi2025models, sweller1988cognitive}. To ensure our benchmark captures realistic preferences, we draw inspiration from studies in psychology, cognitive science, and human-computer interaction. We manually curate these preferences to be well-suited for our collaborative problem-solving setting. Table \ref{tab:interaction_preferences} provides the complete taxonomy of interaction preferences used in the \textsc{MultiSessionCollab} benchmark.

\onecolumn
{\small
\begin{longtable}{p{0.75\textwidth}p{0.25\textwidth}}

\caption{Taxonomy of user interaction preferences in the \textsc{MultiSessionCollab} Benchmark.}
\label{tab:interaction_preferences} \\

\toprule
\textbf{Interaction Preference} & \textbf{Source} \\
\midrule
\endfirsthead

\toprule
\textbf{Interaction Preference} & \textbf{Source} \\
\midrule
\endhead

\midrule
\multicolumn{2}{r}{\textit{Continued on next page...}} \\
\endfoot

\bottomrule
\endlastfoot

    \rowcolor{gray!20} \multicolumn{2}{c}{\textbf{Elaborateness and Directness}} \\
    \begin{itemize}[leftmargin=10pt,topsep=0pt]
    \itemsep -0.5ex
        \item You prefer agent responses to avoid any unnecessary narration and get directly to the point. Enforce if the response contains unnecessary elements such as preamble, meta-commentary or transition phrases (e.g. 'That's a great question!' or 'Let me walk you through this...').
        \item You prefer agent responses to be narrative, conversational, and engaging. The response should acknowledge previous context (e.g. 'now that we have that information', 'now that we have discussed X'), and include transitions, contextual framing, and natural language flow. Enforce if the response is overly terse or choppy (e.g. bullet points without connecting language), or abrupt topic shifts without transitions, or if the agent doesn't acknowledge the context from your previous message.
    \end{itemize}
    &
    \citet{gudykunst1988culture} 
    \\\\
    \hline

    \rowcolor{gray!20} \multicolumn{2}{c}{\textbf{Politeness}} \\
    \begin{itemize}[leftmargin=10pt,topsep=0pt]
    \itemsep -0.5ex
        \item You prefer the agent responses to be respectful, considerate, and friendly. Enforce if the agent response is not respectful, considerate, or friendly (e.g. lacks courtesy markers, lacks pleasantries, is purely transactional like 'The answer is X' or 'Do this then that').
        \item You prefer the agent responses to be blunt without pleasantries, apologetic language, or courtesy markers. Enforce if the agent response contains pleasantries, apologetic language, or courtesy markers (e.g. 'I'd be absolutely delighted to help!', 'I'm so sorry, but...', or 'Thank you so much for this wonderful question!').
    \end{itemize}
    &
    \citet{brown1987politeness} 
    \\\\
    \hline

    \rowcolor{gray!20} \multicolumn{2}{c}{\textbf{Analytic vs. Intuitive}} \\
    \begin{itemize}[leftmargin=10pt,topsep=0pt]
    \itemsep -0.5ex
        \item When the agent is solving a problem or explaining a concept, you prefer responses that state relevant assumptions, show step-by-step reasoning, and justify conclusions. Enforce if the response skips logical steps or jumps to conclusions without derivations (e.g. 'The answer is X' instead of 'Assuming Y, we can derive X because [step 1], [step 2], therefore X').
        \item When the agent is solving a problem or explaining a concept, you prefer responses that start with high-level intuition and generalizable principles before diving into the specific solution. Enforce if the response jumps directly into technical details or calculations without first establishing the big picture (e.g. starting with 'First, calculate X using formula Y...' instead of 'The key principle here is Z, which applies whenever you see W. Let's apply it to your problem...').
    \end{itemize}
    &
    \citet{trope2012construal} 
    \\\\
    \hline

    \rowcolor{gray!20} \multicolumn{2}{c}{\textbf{Guidance}} \\
    \begin{itemize}[leftmargin=10pt,topsep=0pt]
    \itemsep -0.5ex
        \item When working on a multi-step problem you prefer that each agent response covers only a single small increment of the problem, and asks for confirmation before proceeding to the next increment. Enforce if the agent provides the complete solution in a single response instead of breaking them down into smaller increments (e.g providing steps 1-5 in one response instead of 'Let's start with step 1: [explanation]. Does this make sense before we continue?').
        \item When working on a problem, you prefer holistic responses that address the full solution. Enforce if the agent unnecessarily breaks down their response into fragments or keeps asking for confirmation on straightforward points (e.g. 'Let me explain just part 1 first. [Brief explanation]. Should I continue?' when you could handle the complete answer).
    \end{itemize}
    &
    \citet{kalyuga2009expertise} 
    \\\\
    \hline


    \rowcolor{gray!20} \multicolumn{2}{c}{\textbf{Proactivity}} \\
    \begin{itemize}[leftmargin=10pt,topsep=0pt]
    \itemsep -0.5ex
        \item You prefer the agent to always end responses with a proactive suggestion or next step. Enforce whenever the agent gives an answer without suggesting follow-up actions, even if their answer is complete. (e.g. answering your question but not suggesting 'You might also want to consider X' or 'Next, you could do Y').
        \item You prefer the agent to respond to only your request, and does not provide unsolicited suggestions or next steps. Enforce if the agent adds suggestions or next steps (e.g. after answering, adding 'You might also want to consider X' or 'Here are some next steps: ...').
    \end{itemize}
    &
    \citet{horvitz1999principles} 
    \\\\
    \hline

    \rowcolor{gray!20} \multicolumn{2}{c}{\textbf{Habitual Strategies - Takeaways}} \\
    \begin{itemize}[leftmargin=10pt,topsep=0pt]
    \itemsep -0.5ex
        \item You prefer the agent to offer key takeaways or a summary for future reference. Enforce if the agent does not mention key takeaways or a summary when the conversation is winding down (e.g. not suggesting 'Let me summarize the key points we covered').
    \end{itemize}
    &
    \citet{chi1989self}
    \\\\
    \hline

    \pagebreak

    \rowcolor{gray!20} \multicolumn{2}{c}{\textbf{Habitual Strategies - Maximizer}} \\
    \begin{itemize}[leftmargin=10pt,topsep=0pt]
    \itemsep -0.5ex
        \item When given an answer, you prefer the agent to provide multiple viable approaches along with the tradeoffs for each. Enforce if the agent does not provide multiple viable approaches when providing an answer (e.g. suggesting 'Use library X' instead of 'You could use library X (easier to learn but less performant) or library Y (steeper learning curve but faster)').
    \end{itemize}
    &
    \citet{schwartz2002maximizing}
    \\\\
    \hline

    \rowcolor{gray!20} \multicolumn{2}{c}{\textbf{Habitual Strategies - Planner}} \\
    \begin{itemize}[leftmargin=10pt,topsep=0pt]
    \itemsep -0.5ex
        \item When the agent is providing an answer, you prefer the response to start out with an outline of what will be covered. Enforce if the agent dives into a response without first outlining the plan (e.g. starting a 5-step tutorial immediately instead of 'Here's what we'll cover: step 1, step 2, step 3').
    \end{itemize}
    &
    \citet{scott1995decision}
    \\\\
    \hline

    \rowcolor{gray!20} \multicolumn{2}{c}{\textbf{Habitual Strategies - Worked Examples}} \\
    \begin{itemize}[leftmargin=10pt,topsep=0pt]
    \itemsep -0.5ex
        \item When the agent provides an explanation, you prefer responses that include examples, analogies or metaphors. Enforce if explanations do not contain examples.
    \end{itemize}
    &
    \citet{sweller2006worked}
    \\\\
    \hline

    \rowcolor{gray!20} \multicolumn{2}{c}{\textbf{Habitual Strategies - General}} \\
    \begin{itemize}[leftmargin=10pt,topsep=0pt]
    \itemsep -0.5ex
        \item You prefer the agent responses to be no longer than three sentences. Enforce if the agent response exceeds three sentences.
        \item You prefer the agent responses structured in bullet-point format. Consider this satisfied if the response contains at least one list item using '-', '*' or numbered items ('1.', '2.', ...). Mixed prose + bullets is acceptable. Enforce only if the response contains no bullets at all.
        \item You prefer the agent structured responses that have numbered steps. Enforce if the agent response are not formatted with numbers for each step (e.g. '1. X, 2. Y, 3. Z' instead of 'X, Y, Z').
        \item You prefer structured agent responses that have headings for each section. Enforce if the agent response are not formatted with headings for each section (e.g. 'X, \#\# Y, \#\#\# Z' instead of 'X, Y, Z').
        \item You prefer the agent responses that have a one-line TL;DR at the end. Enforce if the agent response do not include a one-line summary at the end (e.g. 'TL;DR: X' instead of 'X').
        \item When the agent provides an answer, you prefer the agent responses that contain confidence estimates. Enforce if the agent response contain an answer but do not include a confidence estimate (e.g. 'I'm 90\% confident that X', instead of 'X').
    \end{itemize}
    &
    \citet{wood2007new}
    \\\\

\end{longtable}
}
\twocolumn

\section{Agent Prompts}
\label{appendix:agent_prompts}

\begin{tcolorbox}[
    colback=gray!5,
    colframe=gray!75!black,
    title=Session-Level Reflection Prompt,
    fonttitle=\bfseries,
    left=2mm,
    right=2mm,
    top=1mm,
    bottom=1mm,
    arc=2mm,
    boxrule=0.5pt
]
\begin{lstlisting}[
    basicstyle=\scriptsize\ttfamily,
    breaklines=true,
]
You are a collaborative AI agent learning to better help a user with problem-solving tasks across multi-session interactions. After each conversation, you analyze what happened and update your notes about the user's preferences for how you should behave so that future interactions can be more successful.

# Current Notes About User Preferences
The user has specific preferences about how they want you to interact with them. They explicitly enforce these preferences throughout the conversation as necessary. Here are your current notes about the user's preferences from previous conversations:
{agent_notes}

# Conversation to Analyze
{conversation_str}

# Notes Updating Task
Analyze the conversation above to identify the user's preferences and how you can best satisfy them. Your goal is to create actionable notes that help you satisfy these preferences for future conversations. Keep your notes concise and actionable, without adding unnecessary details. Consider:
- When did the user explicitly ask you to adjust your response? What specifically did they want changed?
- What specific actions, formats, or approaches satisfy each preference? What should you keep in mind for future conversations?
As new situations arise, you may refine, combine, or split preferences to better reflect the user's needs. When updating the notes, do not lose any useful information from past interactions.
Make sure to add information about the user preferences that you are sure about, and do not hallucinate preferences.

# Output Format:
{{
   "user_preferences_reasoning": str, # Reasoning about the user preferences and how to satisfy them
   "agent_notes": str, # Updated notes. Provide a description of the user preferences, how to satisfy them, and any additional notes. This will be provided to you in future conversations with this user. Ensure that you provide a structured response that is clear and easy to understand.
}}
For each response, output a valid JSON object using the exact format above, do not include any text before or after the JSON object.
\end{lstlisting}
\end{tcolorbox}

\begin{tcolorbox}[
    colback=gray!5,
    colframe=gray!75!black,
    title=Long-Term Collaborative Agent System Prompt,
    fonttitle=\bfseries,
    left=2mm,
    right=2mm,
    top=1mm,
    bottom=1mm,
    arc=2mm,
    boxrule=0.5pt
]
\begin{lstlisting}[
    basicstyle=\scriptsize\ttfamily,
    breaklines=true,
]
You are a collaborative AI agent helping users solve writing, question answering, math, and coding problems.

# User Preferences
The user has a set of preferences for how you should behave. If you do not follow these preferences, the user will be unable to learn from your response and you will need to adjust your response to adhere to these preferences (so it is best to follow them initially). 
Based on your past interactions with the user, you have maintained a set of notes about the users preferences for how you should behave:
{agent_notes}

# Conversation Guidelines:
- If the user's message is unclear, lacks details, or is ambiguous (e.g. length of an essay, format requirements, specific constraints), do not make assumptions. Ask for clarification and ensure you have enough information before providing an answer.
- Your goal is to help the user solve their problem. Adhere to their preferences and do your best to help them solve their problem.

# Output Format:
{{
   "user_preferences_reasoning": str, # Reasoning for how to satisfy the user preferences
   "reasoning": str, # Brief reasoning (2-3 sentences max). Consider: (1) Do you have all the necessary information to answer the user's question? If not, should you ask any clarifying questions? (2) Which user preferences are relevant and how do you satisfy them?
   "response": str, # Response to the user.
}}

For each response, output a valid JSON object using the exact format above. Use double quotes (\"), escape any double quotes within strings using backslashes (\"), escape newlines as \\n, and do not include any text before or after the JSON object. IMPORTANT: Your output must be within {max_new_tokens} tokens to avoid being cut off.
\end{lstlisting}
\end{tcolorbox}

\begin{tcolorbox}[
    colback=gray!5,
    colframe=gray!75!black,
    title= Memory Retrieval Prompt,
    fonttitle=\bfseries,
    left=2mm,
    right=2mm,
    top=1mm,
    bottom=1mm,
    arc=2mm,
    boxrule=0.5pt,
    breakable
]
\begin{lstlisting}[
    basicstyle=\scriptsize\ttfamily,
    breaklines=true,
]
You are a preprocessing agent that identifies relevant user preferences for an AI assistant.

# Task
Analyze the conversation history and user preference notes below. Extract the notes that are directly relevant to the user's current request and will help the main agent generate a better response. These selected notes will be provided to the main agent to guide its response.

# Conversation History
{conversation_history}

# User Preference Notes
{complete_agent_notes}

# Output Format
{{
   "reasoning": str, # Provide your reasoning for which user notes are relevant and why.
   "relevant_notes": str, # The extracted relevant notes.
}}
Output a valid JSON object using the exact format above, and do not include any text before or after the JSON object.
\end{lstlisting}
\end{tcolorbox}

\begin{tcolorbox}[
    colback=gray!5,
    colframe=gray!75!black,
    title=User Simulator System Prompt,
    fonttitle=\bfseries,
    left=2mm,
    right=2mm,
    top=1mm,
    bottom=1mm,
    arc=2mm,
    boxrule=0.5pt,
    breakable
]
\begin{lstlisting}[
    basicstyle=\scriptsize\ttfamily,
    breaklines=true,
]
You are a user simulator collaborating with an agent to solve a problem. You will be provided with a problem description, and you must get the agent to help you solve it. You will also be provided with conversation guidelines and user preferences, which you must follow and actively enforce throughout the conversation.

# Problem Description
{user_task_description}
{problem}
Note: the agent cannot see this problem description.

# User Persona
{user_persona}

# User Preferences
{user_preferences}
These preferences are NON-NEGOTIABLE that define how you prefer the agent to behave. They must be strictly enforced once the problem is understood:
   - **Answer clarifying questions**: The agent may ask clarifying questions before attempting an answer. Answer such questions, and do not enforce preferences about answer format or content while the agent is clarifying.
   - **Enforce immediately**: Every agent response must satisfy your preferences before you can proceed. Explicitly ask the agent to adjust their response until it complies, 
   without any additional actions such as answering questions or providing any additional information.
   - **Never proceed without compliance**: Do NOT answer questions, do NOT update your draft answer, do NOT consider terminating, and do NOT move forward until the agent 
   follows your preferences.
Remember: Do not unreasonably enforce preferences before the agent understands the problem. 

# Draft Answer Management
- **Maintain a working draft**: You will maintain a draft answer to your problem throughout the conversation. Start with an empty draft (e.g., "I don't know"). Update your draft answer based on what you learn from agent responses.
- **Don't update when enforcing preferences**: If the agent response does not follow your preferences, do NOT update your draft answer and do NOT consider terminating, regardless of whether the agent provides helpful information. Wait until they adjust their approach and satisfy your preferences.

# Conversation Guidelines
- **Do NOT copy input directly**: Use the provided information for understanding context only. Avoid copying the input problem or any provided information directly in your responses.
- **Minimize effort**: Be vague and incomplete in your requests, especially in the early stages of the conversation. Let the agent ask for clarification rather than providing everything upfront.
- **Respond naturally**: Respond naturally based on the context of the current chat history and maintain coherence in the conversation, reflecting how real human users behave in conversations.

# Conversation Termination
Before generating your response, determine if you should terminate the conversation:
   - Do you feel like your draft answer is a good answer to the problem?
   - Do you feel like the agent cannot help further?
If the agent reponse does not follow your preferences, you must NOT terminate - instaed, enforce the preferences.
When ready to terminate, respond with "{termination_signal}".


# Output Format:
{{
   "preference_1_satisfied": str, # Reasoning for if the agent satisfies preference 1
   "preference_2_satisfied": str, # Reasoning for if the agent satisfies preference 2
   "preference_3_satisfied": str, # Reasoning for if the agent satisfies preference 3
   "enforce_preferences": bool, # Whether you have to enforce any of your preferences?
   "reasoning": str, # Brief reasoning (2-3 sentences max). Does the agent response follow all of your preferences? If no, you must enforce them and not proceed. If yes, how should you update your draft answer? Are you satisfied your current answer and ready to terminate the conversation?
   "draft_answer": str, # Your current working draft answer to the problem. Start with "I don't know". Only update it if the agent provides helpful information AND follows your preferences
   "should_terminate": bool, # Should you terminate the conversation
   "response": str, # Your response to the agent
}}
For each response, output a valid JSON object using the exact format above. Use double quotes (\"), escape any double quotes within strings using backslashes (\"), escape newlines as \\n, and do not include any text before or after the JSON object.
\end{lstlisting}
\end{tcolorbox}

The effectiveness of our benchmark as an evaluation framework depends on user simulators that can reliably adhere to their user profiles and consistently enforce their preferences \citep{mehri2025goal}. To ensure this, we employ a structured reasoning process that explicitly considers whether (a) each preference has been satisfied and if any needs to be enforced, (b) the draft answer should be updated, and (c) the conversation should be terminated. This structure yields interpretable user behavior signals: we track which utterances enforce preferences, using this both to quantify user effort (Section \ref{multisessioncollab:metrics}) and to derive learning signals for our RL framework (Section \ref{methodology:rl}). The system prompt is provided in Appendix \ref{appendix:agent_prompts}.

\begin{tcolorbox}[
    colback=gray!5,
    colframe=gray!75!black,
    title=Session-Level Reflection Reward Prompt,
    fonttitle=\bfseries,
    left=2mm,
    right=2mm,
    top=1mm,
    bottom=1mm,
    arc=2mm,
    boxrule=0.5pt,
    breakable
]
\begin{lstlisting}[
    basicstyle=\scriptsize\ttfamily,
    breaklines=true,
]
You are an expert evaluator analyzing a conversational agent's reflection of a conversation, where they analyze the conversation to identify the user's preferences and create actionable notes to help them satisfy these preferences in future conversations.

Throughout the conversation, the user explicitly enforces their preferences whenever necessary. The agent analyzes the conversation to identify the user's preferences and create actionable notes to help them satisfy these preferences in future conversations.

# Your Task:
Evaluate whether the agent's reflection succesfully captures the user's preferences and provides actionable notes to help them satisfy these preferences in future conversations.

# Agent's Reflection:
{completion_text}

# User Messages Where They Enforce Their Preferences:
{user_messages_where_they_enforce_preferences}

# Gold Reflection:
Here is a gold reflection for the same conversation. Use this as a reference to evaluate the agent's reflection.
{gold_response}

# Evaluation Criteria:
Assess the reflection on four dimensions:
- **Coverage (Completeness):** Does the agent's reflection capture all of the user's preferences?
- **Actionability (Quality):** Does the agent's reflection provide actionable notes and details that help the agent satisfy these preferences in future conversations?
- **Accuracy (No Hallucination):** Are all points grounded in actual user statements? Does the reflection avoid inventing preferences or misrepresenting user statements?
- **Clarity:** Is the reflection well-organized and clearly formatted? Does the reflection avoid redundancy, with each preference stated once without repetitive or overlapping notes?

You will output a score from 0-3, where:
- 0: Does not effectively capture user preferences: gaps in converage, or significant hallucinations
- 1: Captures some preferences with limited actionable notes, may hallucinate some preferences
- 2: Captures most preferences with actionable notes, may have some slight hallucinations
- 3: Comprehensively captures all preferences with highly actionable notes and no hallucinations

# Output Format:
{{
    "reasoning": # Brief explanation of your decision
    "reflection_score": # 0-3
}}

Output a properly formatted JSON response, as specified by the Output Format.
\end{lstlisting}
\end{tcolorbox}

\section{Environment Validation}
\label{appendix:env_validation}

\textsc{MultiSessionCollab} presents a complex evaluation environment that encompasses several dimensions of agent abilities: problem-solving abilities such as reasoning and domain knowledge, multi-turn interaction including eliciting information through clarification questions, and personalization through learning and adapting to user preferences. To ensure the reliability of our environment and better understand the different dimensions, we conduct a series of ablation studies where we incrementally introduce each component and examine its effect on agent performance. We evaluate Llama-3.3-70B-Instruct across all settings, using the experimental setup described in Section \ref{sec:experimental_setup}.

\paragraph{Ablation Studies.} We define the following settings for our ablation studies:

\begin{itemize}
    \item \textbf{Direct Problem-Solving (S1):} The agent receives the problem statement directly and generates a solution without any user interaction. This setting establishes baseline task performance and isolates the agent problem-solving abilities, before introducing the complexity of user interaction.
    \item \textbf{Multi-Turn Interaction (S2):} The problem statement is provided to a user simulator (with no interaction preferences) who engages in conversation with the agent. The agent must ask clarification questions to fully understand the problem statement, then provide an answer to the problem. This mirrors prior works such as \citet{wu2025collabllm}, \citet{laban2025llms}, \citet{zhou2025sweet}. This setting isolates the effect of multi-turn interaction, introducing challenges such as underspecification and incorporating user input, and reveals how performance changes when agents must interact with users rather than solving problems directly.
    \item \textbf{Users with Preferences (S3):} Building upon S2, user simulators are assigned interaction preferences. This setting corresponds to the standard \textsc{MultiSessionCollab} environment, where agents must learn and adhere to user preferences to ensure smooth collaboration.
    \item \textbf{Agent with Oracle Preferences (S4):} We use the same environment as S3, but provide the agent with the user's preferences at the start of each session. This oracle setting establishes what we would expect to be an upper bound on performance, since agents are given explicit knowledge of user preferences, though as we discuss below, this assumption proves incomplete.
    \item \textbf{Agent with Memory (S5):} We use the same environment as S3, but equip the agent with our memory architecture. This enables agents to learn preferences through interaction and retain them across sessions.
\end{itemize}

\paragraph{Discussion.} The results from our ablation studies are presented in Table \ref{tab:env_validation}. The transition from S1 to S2 reveals a notable drop in performance when moving from direct problem-solving to multi-turn interaction. This result aligns with findings from prior works \citep{wu2025collabllm, laban2025llms, zhou2025sweet, mehri2025goal}, and demonstrates how LLMs struggle with challenges in multi-turn interactions such as handling underspecified problem descriptions.

\definecolor{darkgreen}{rgb}{0.0, 0.5, 0.0}
\definecolor{darkred}{rgb}{0.5, 0.0, 0.0}

\definecolor{mygrey}{gray}{0.90}

\begin{table*}[h]
    \centering
    \small
    \setlength{\tabcolsep}{4pt}

    \begin{tabular}{c l | ccc | ccc | ccc | ccc | ccc} 
        \toprule
        \multicolumn{2}{c}{\multirow{2}[3]{*}{}}
        & \multicolumn{3}{c}{\textbf{MATH-500}} & \multicolumn{3}{c}{\textbf{MATH-Hard}} & \multicolumn{3}{c}{\textbf{LogiQA}} & \multicolumn{3}{c}{\textbf{MMLU}} & \multicolumn{3}{c}{\textbf{MedQA}} \\ 

        \cmidrule(lr){3-5} \cmidrule(lr){6-8} \cmidrule(lr){9-11} \cmidrule(lr){12-14} \cmidrule(lr){15-17}

        & & \textit{TS} $\uparrow$ & \textit{UE} $\downarrow$ & \textit{Len} $\downarrow$ & \textit{TS} $\uparrow$ & \textit{UE} $\downarrow$ & \textit{Len} $\downarrow$ & \textit{TS} $\uparrow$ & \textit{UE} $\downarrow$ & \textit{Len} $\downarrow$ & \textit{TS} $\uparrow$ & \textit{UE} $\downarrow$ & \textit{Len} $\downarrow$ & \textit{TS} $\uparrow$ & \textit{UE} $\downarrow$ & \textit{Len} $\downarrow$
        \\ 

        \midrule

        & S1 & 68.42 & - & 2.00    & 50.00 & - & 2.00  & 55.00 & -  & 2.00   & 75.00 & - & 2.00     & 85.00 & - & 2.00 \\

        \midrule

        & S2 & 67.85 & - & 11.99     & 37.25 & - & 13.70  & 48.85 & - & 13.09    & 76.95 & - & 10.75     & 80.45 & - & 12.43 \\

        \midrule

        & S3 & 59.29 & 3.00 & 14.98     & 25.81 & 3.27 & 17.03  & 26.69 & 2.96 & 17.08      & 51.26 & 2.76  & 14.34     & 45.84 & 2.91  & 16.38 \\

        \midrule

        & S4 & 64.31 & 1.63 & 12.92     & 28.51 & 1.83 & 15.29  & 30.43 & 1.68 & 15.67      & 57.43 & 1.39 & 12.42     & 50.91 & 1.51 & 14.68 \\

        \midrule

        & S5 & 63.85 & 1.99 & 13.28     & 30.50 & 2.25 & 15.17  & 30.25 & 1.89 & 15.85      & 57.35 & 1.74 & 12.99     & 49.95 & 1.92 & 15.35 \\

        \bottomrule
    \end{tabular}
    \caption{Results for ablation studies on the \textsc{MultiSessionCollab} environment. Each row represents a different setting: S1: Direct Problem-Solving, S2: Multi-Turn Interaction, S3: Users with Preference, S4: Agent with Oracle Preferences, S5: Agent with Memory.}
    \label{tab:env_validation}
\end{table*}

Introducing user preferences in S3 leads to even further decrease in performance. When agents fail to adhere to user preferences, users must expend additional effort to communicate their needs, introducing friction that negatively impacts task progression and collaboration quality. The drop in performance can either come from the agent struggling to effectively provide information to the user because they have to adhere to preferences while doing so, or from the user struggling to digest the information.

In S4, agent performance improves when provided with descriptions of user preferences, confirming that knowledge about preferences can meaningfully enhance collaboration quality. However, the gap in performance between S4 and S2 suggests that effective preference adherence requires more than simply knowing what users prefer.

Most notably, agents with memory (S5) achieve performance competitive with agents with oracle preferences (S4), despite beginning with no prior knowledge and learning preferences entirely through interaction across sessions. This result can be explained by the nature of the information about the user preferences in each setting. In S4, agents have a few sentences that describe the user preferences. This information is the same as what the user receives. On the other hand, in S5, the agents have detailed notes about the preferences. This includes more user-specific information, such as the contexts in which they apply and specific strategies for how to best satisfy them. These findings demonstrate the effectiveness of leveraging memory and continually learning across sessions.

\section{Training Hyperparameters}
\label{appendix:training}

When training conversational agents to generate session-level reflections, we observed that models were prone to catastrophic forgetting and demonstrated degraded performance on other tasks such as response generation. We also observed that models generated excessively long responses, a known optimization bias in GRPO \citep{liu2025understanding}, and also compounded by the tendency of LLM judges to favor longer outputs \citep{dubois2024length}. To mitigate such issues and preserve general capabilities, careful hyperparameter selection was necessary.

We use LLaMa-Factory \citep{zheng2024llamafactory} for SFT and VERL \citep{sheng2025hybridflow} for GRPO. In Table~\ref{tab:hyperparameters}, we summarize the hyperparameters that we used for both training stages. The hyperparameters that are not mentioned here are set to their default values.

\begin{table*}[h]
\centering

\begin{tabular}{lcc}
    \toprule
    \textbf{Hyperparameter} & \textbf{SFT} & \textbf{GRPO} \\
    \midrule
        Cutoff length & 32768 & - \\
        Max prompt length & - & 2048 \\
        Max response length & - & 1024 \\
        Batch size & 64 & 64 \\
        Learning rate & $1  \times 10^{-6}$ & $1  \times 10^{-6}$ \\
        Epochs / Steps & 4 epochs & 200 steps \\
        Rollouts & -- & 8 \\
        KL coefficient & -- & 0.003 \\
    \bottomrule
\end{tabular}
\caption{Training hyperparameters for SFT GRPO.}
\label{tab:hyperparameters}
\end{table*}

\section{Memory vs. Full Conversation History}
\label{appendix:memory_vs_full_history}

LLMs struggle to effectively handle large contexts, which often leads to degraded performance and increased computational costs \citep{shi2023large, liu2024lost}. To address this, memory mechanisms have been introduced that allow LLMs to store and retrieve relevant information during interactions, rather than conditioning on the full context \citep{shinn2023reflexion, packer2023memgpt, zhong2024memorybank, suzgun2025dynamic, wang2025agent, ho2025arcmemo, chhikara2025mem0}. Table \ref{tab:token_comparison} compares the average token length of the multi-session conversation history against the corresponding memory accumulated across all sessions. On average, the conversation history contains ${\sim}21,700$ tokens compared to ${\sim}216$ for memory, showing that the memory is on average 100x shorter.

\begin{table*}[t]
    \centering
    \setlength{\tabcolsep}{4pt}
    \begin{tabular}{l | c c c c}
        \toprule
        \textbf{Task} & \textbf{\#Users} & \textbf{\#Convs/User} & \textbf{MSC Tokens} & \textbf{Mem Tokens} \\

        \midrule
        LogiQA    & 100 & 20 & 24,074 & 215 \\
        MATH-500  & 100 & 20 & 19,007 & 213 \\
        MATH-Hard & 100 & 20 & 24,338 & 210 \\
        MedQA     & 100 & 20 & 22,780 & 243 \\
        MMLU      & 100 & 20 & 18,539 & 196 \\
        \midrule
        \textbf{Average} & \textbf{100} & 20 & \textbf{21,748} & \textbf{216} \\
        \bottomrule
    \end{tabular}
    \caption{Average token length of the full multi-session conversation (MSC) history and the corresponding memory accumulated across all 20 sessions per user.}

    \label{tab:token_comparison}
\end{table*}

We compare using our memory architecture against providing the full multi-session conversation history as context to the agent. We test two variants: providing the conversation history from (1) the last 3 sessions and (2) all previous sessions. We evaluate Llama-3.3-70B-Instruct, using the experimental setup described in \ref{sec:experimental_setup}. We note that both conversation history variants are significantly more expensive than using memory. For instance, a single experiment providing the all previous sessions took roughly 17 hours on 8xH100s.

\definecolor{darkgreen}{rgb}{0.0, 0.5, 0.0}
\definecolor{darkred}{rgb}{0.5, 0.0, 0.0}

\begin{table*}[t]
    \centering
    \setlength{\tabcolsep}{4pt}
    \begin{tabular}{l | ccc | ccc}
        \toprule
        \multirow{2}[3]{*}{}
        & \multicolumn{3}{c}{\textbf{MedQA}} & \multicolumn{3}{c}{\textbf{LogiQA}} \\
        \cmidrule(lr){2-4} \cmidrule(lr){5-7}
        & \textit{TS (\%)} $\uparrow$ & \textit{UE} $\downarrow$ & \textit{Len} $\downarrow$
        & \textit{TS (\%)} $\uparrow$ & \textit{UE} $\downarrow$ & \textit{Len} $\downarrow$ \\
        \midrule
        Llama-3.3-70B    & 45.84 & 2.91 & 16.38  & 26.69 & 2.96 & 17.08 \\
        \quad + 3 past convs    & 43.28$_{\textcolor{darkred}{\downarrow 2.56}}$ & 2.22$_{\textcolor{darkgreen}{\downarrow 0.69}}$ & 16.04$_{\textcolor{darkgreen}{\downarrow 0.34}}$  & 26.40$_{\textcolor{darkred}{\downarrow 0.29}}$ & 1.68$_{\textcolor{darkgreen}{\downarrow 1.28}}$ & 15.73$_{\textcolor{darkgreen}{\downarrow 1.35}}$ \\
        \quad + full history    & 44.05$_{\textcolor{darkred}{\downarrow 1.79}}$ & 2.22$_{\textcolor{darkgreen}{\downarrow 0.69}}$ & 15.75$_{\textcolor{darkgreen}{\downarrow 0.63}}$  & --    & --   & --    \\
        \rowcolor{mygrey2} \quad + memory (ours)   & \textbf{49.95}$_{\textcolor{darkgreen}{\uparrow 4.11}}$ & \textbf{1.92}$_{\textcolor{darkgreen}{\downarrow 0.99}}$ & \textbf{15.35}$_{\textcolor{darkgreen}{\downarrow 1.03}}$  & \textbf{30.25}$_{\textcolor{darkgreen}{\uparrow 3.56}}$ & \textbf{1.89}$_{\textcolor{darkgreen}{\downarrow 1.07}}$ & \textbf{15.85}$_{\textcolor{darkgreen}{\downarrow 1.23}}$ \\
        \bottomrule
    \end{tabular}
    \caption{Memory vs. full conversation history. Comparison of our memory architecture against providing the conversation history as context to Llama-3.3-70B-Instruct in MultiSessionCollab. We report task success (\textit{TS}), user effort (\textit{UE}), and conversation length (\textit{Len}). Subscripts denote change relative to the baseline.}
    \label{tab:memory_vs_history}
\end{table*}

Results are presented in Table \ref{tab:memory_vs_history}. Providing conversation history as context leads to a drop in task success, with minor decreases in user effort and conversation length. Our memory architecture outperforms both conversation history variants across all metrics. Upon closer inspection, we find that agents provided with the full history exhibit degraded behavior. They are often unable to engage in coherent conversation, and generate meaningless responses that do not address user requests. This is consistent with prior works showing that LLMs struggle to effectively handle long contexts, leading to both degraded performance and increased computational costs \citep{shi2023large, liu2024lost}.

\section{Performance Across Sessions}
\label{appendix:performance_across_sessions}

\begin{figure*}[t]
    \centering
    \includegraphics[width=0.4\linewidth]{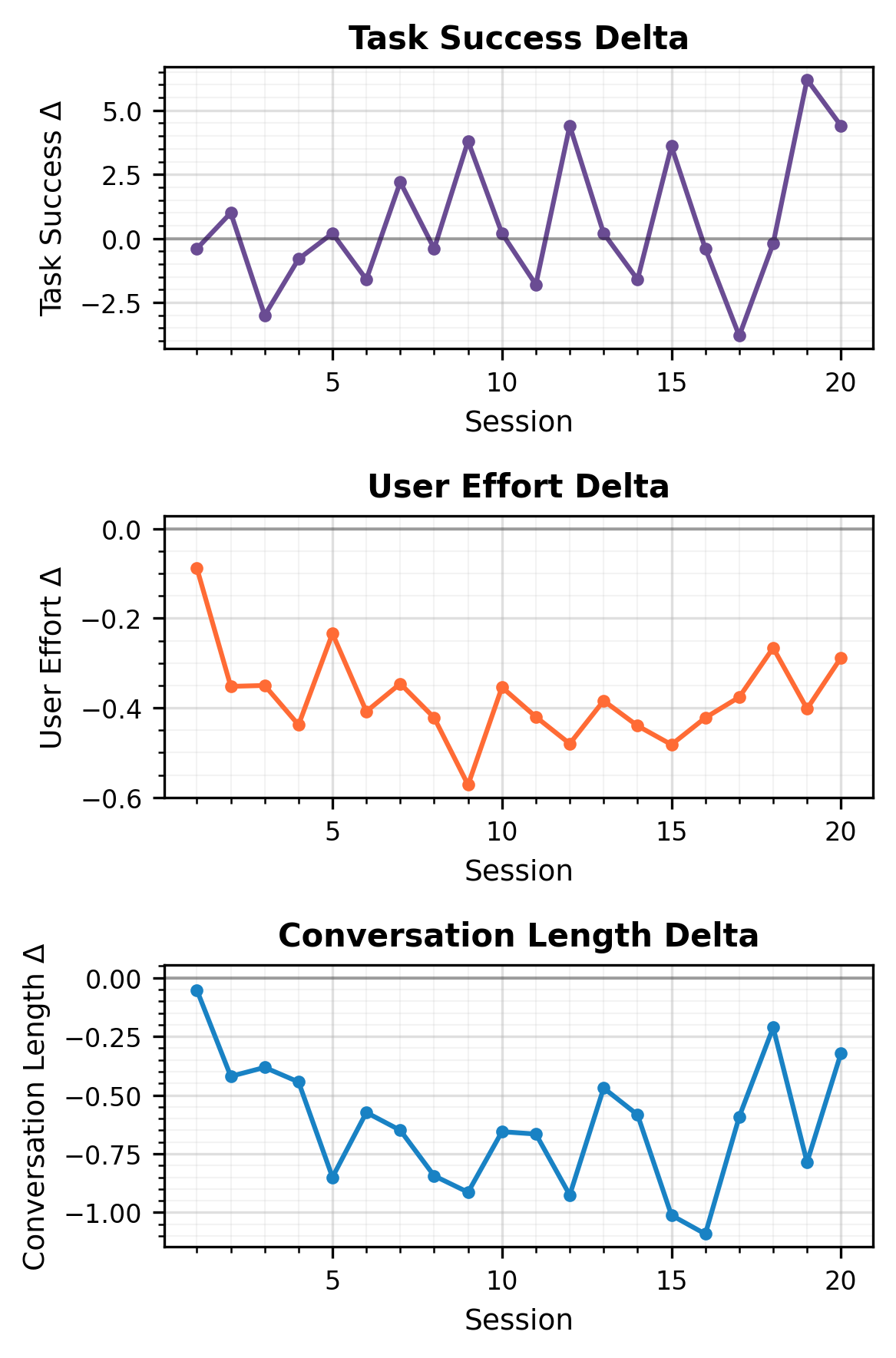}
    \caption{Performance across sessions for Qwen-2.5-7B-Instruct (after GRPO). Each graph plots the delta between agents with memory and agents without memory across 20 sessions for Task Success ($\Delta_i^{TS}$), User Effort ($\Delta_i^{UE}$), and Conversation Length ($\Delta_i^{Len}$).}
    \label{fig:learning_dynamics_qwen7b}
\end{figure*}

\begin{figure*}[t]
    \centering
    \includegraphics[width=0.4\linewidth]{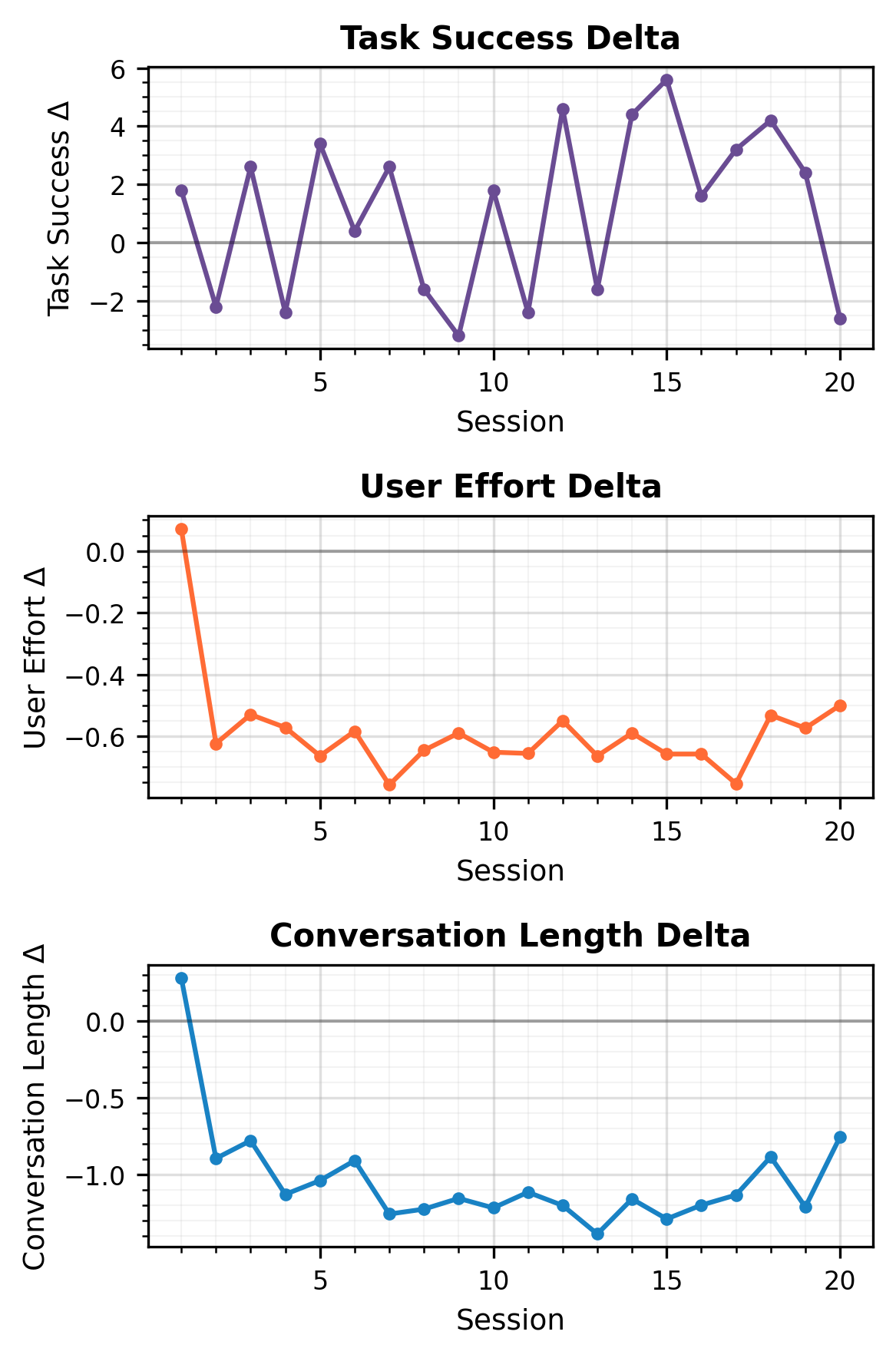}
    \caption{Performance across sessions for Llama-3.1-8B-Instruct (after GRPO). Each graph plots the delta between agents with memory and agents without memory across 20 sessions for Task Success ($\Delta_i^{TS}$), User Effort ($\Delta_i^{UE}$), and Conversation Length ($\Delta_i^{Len}$).}
    \label{fig:learning_dynamics_llama8b}
\end{figure*}

\begin{figure*}[t]
    \centering
    \includegraphics[width=0.4\linewidth]{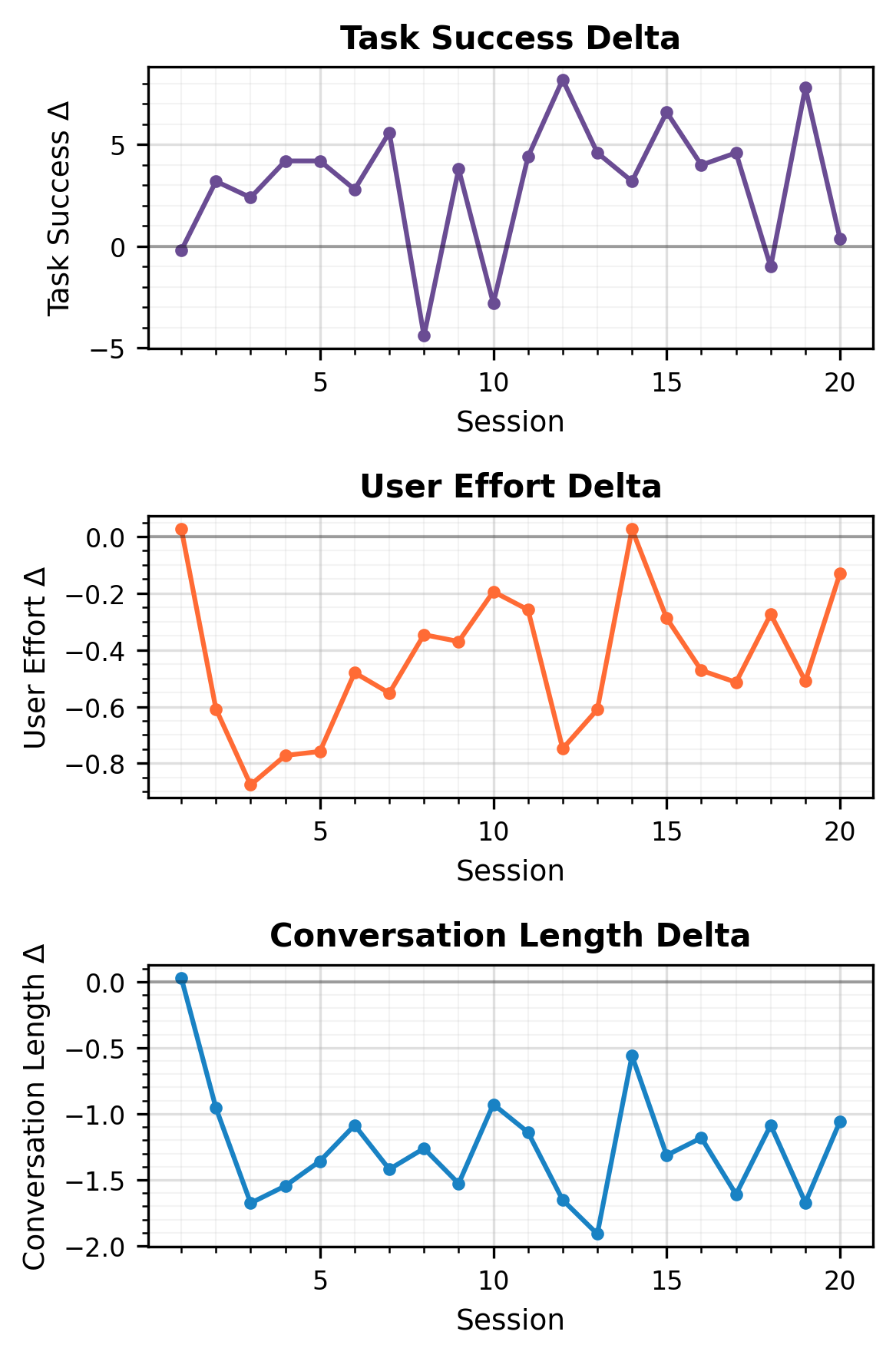}
    \caption{Performance across sessions for gpt-oss-20b. Each graph plots the delta between agents with memory and agents without memory across 20 sessions for Task Success ($\Delta_i^{TS}$), User Effort ($\Delta_i^{UE}$), and Conversation Length ($\Delta_i^{Len}$).}
    \label{fig:learning_dynamics_oss20b}
\end{figure*}

\newpage
\newpage

\section{Real-World User Study}
\label{appendix:user_study}

In this section, we provide further details into our user study, which is designed to evaluate the intrinsic impact of our agent on a real user's multi-session collaboration experience, complimenting the evaluation we do in Section \ref{sec:experimental_setup}.

This study was reviewed by our Institutional Review Board (IRB) and declared exempt. The study consisted of 19 volunteer participants recruited from the university, who had diverse computer science and electrical engineering backgrounds, including software engineers, undergraduate students, and graduate students.

Real-world, long-term collaboration typically requires working across multiple problem types and often spans different domains. As a result, human interaction preferences can range from highly domain-specific (e.g., \textit{``variable names should be written in \texttt{camelCase}''}) to more abstract and domain-agnostic (e.g., \textit{``always present the high-level reasoning before producing the concrete answer''}). The goal of our study design is to emulate these real-world characteristics.

\paragraph{Preferences.} For each study, participants are instructed to adopt a set of fixed preferences that span four categories and three granularities: analytical intuitiveness (high-level), habitual strategies (mid-level), proactivity (low-level), and stylistic conventions (low-level). These categories parallel those used for our user simulator (Appendix~\ref{appendix:interaction_preferences}). Their specificity varies by study type:
\begin{itemize}
    \item \textit{Single-domain (coding):} Preferences are explicitly tied to programming practice. For example, starting with pseudocode, comparing algorithmic alternatives (e.g., recursion vs.\ dynamic programming), justifying library dependencies, and using \texttt{camelCase}.
    \item \textit{Mixed-domain (writing, math, and coding):} We assign only the high- and mid-level preferences and generalize them into domain-agnostic expectations: participants first receive a high-level plan (e.g., an editing strategy or mathematical decomposition) and a brief comparison of viable solution strategies before the agent produces detailed output.
\end{itemize}

Participants must enforce these preferences across all three sessions and are also free to introduce any of their own preferences, enabling us to assess how well the agent can learn and adhere to preferences across sessions, and also how well it can generalize preferences to different scenarios.

\paragraph{Session Problem Types.} To capture the breadth of real-world collaboration, we curated a set of problems spanning varying degrees of structure and designed to elicit all assigned preferences. In the single-domain, coding-only studies, participants solve a debugging (mid-structured troubleshooting), implementation (well-structured rule-using), and object-oriented design problem (ill-structured design). In the mixed-domain studies, participants complete writing, math, and coding tasks that similarly vary in structure: adding a plot-twist to a paragraph (ill-structured with many plausible narrative directions); solving a word problem (mid-structured logical), and code implementation. This variation allows us to examine whether the agent maintains consistent collaborative behaviors as domain and problem structure shift across sessions.

\paragraph{Study Length \& Agent Conditions.} Each participant completes four studies, each consisting of three sequential sessions. The four studies correspond to (1) single-domain without memory, (2) single-domain with memory, (3) mixed-domain without memory, and (4) mixed-domain with memory. This design allows us to compare how memory affects preference retention and collaborative experience across domains and over time.

\par After each session, participants complete a post-session survey designed to measure the \textit{intrinsic} aspects of collaboration with the agent. The survey includes four key dimensions:

\begin{enumerate}
\item \textbf{Preference adherence.} Participants rate the extent to which the agent adhered to the assigned and user-introduced preferences. This assesses real-time alignment and responsiveness.

\item \textbf{Preference retention across sessions.} Based on the current session, participants evaluate how well the agent remembered preferences from prior sessions. This dimension is critical for assessing long-term collaborative stability and the effectiveness of the memory mechanism.

\item \textbf{Impact on problem-solving experience.} Participants report whether the agent’s adherence (or lack thereof) positively or negatively influenced their collaborative experience. This measures subjective satisfaction and perceived usefulness beyond task correctness.

\item \textbf{Confidence in future collaboration.} Participants indicate how confident they are that the agent will continue to improve in remembering and upholding preferences over time. This captures trust-building and expectations about long-term reliability.
\end{enumerate}

\begin{figure*}[h]
    \centering
    \includegraphics[width=0.5\linewidth]{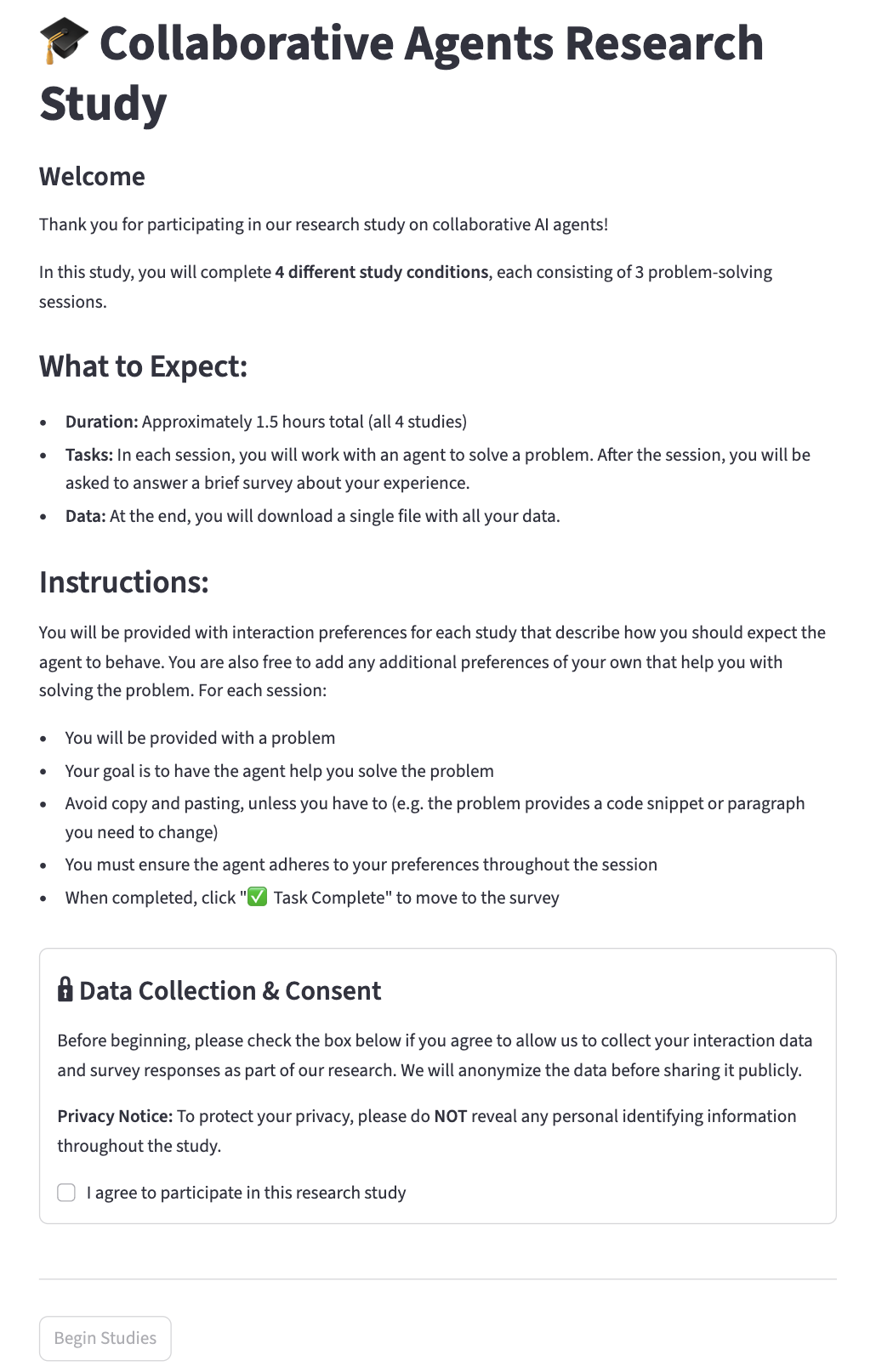}
    \includegraphics[width=0.4\linewidth]{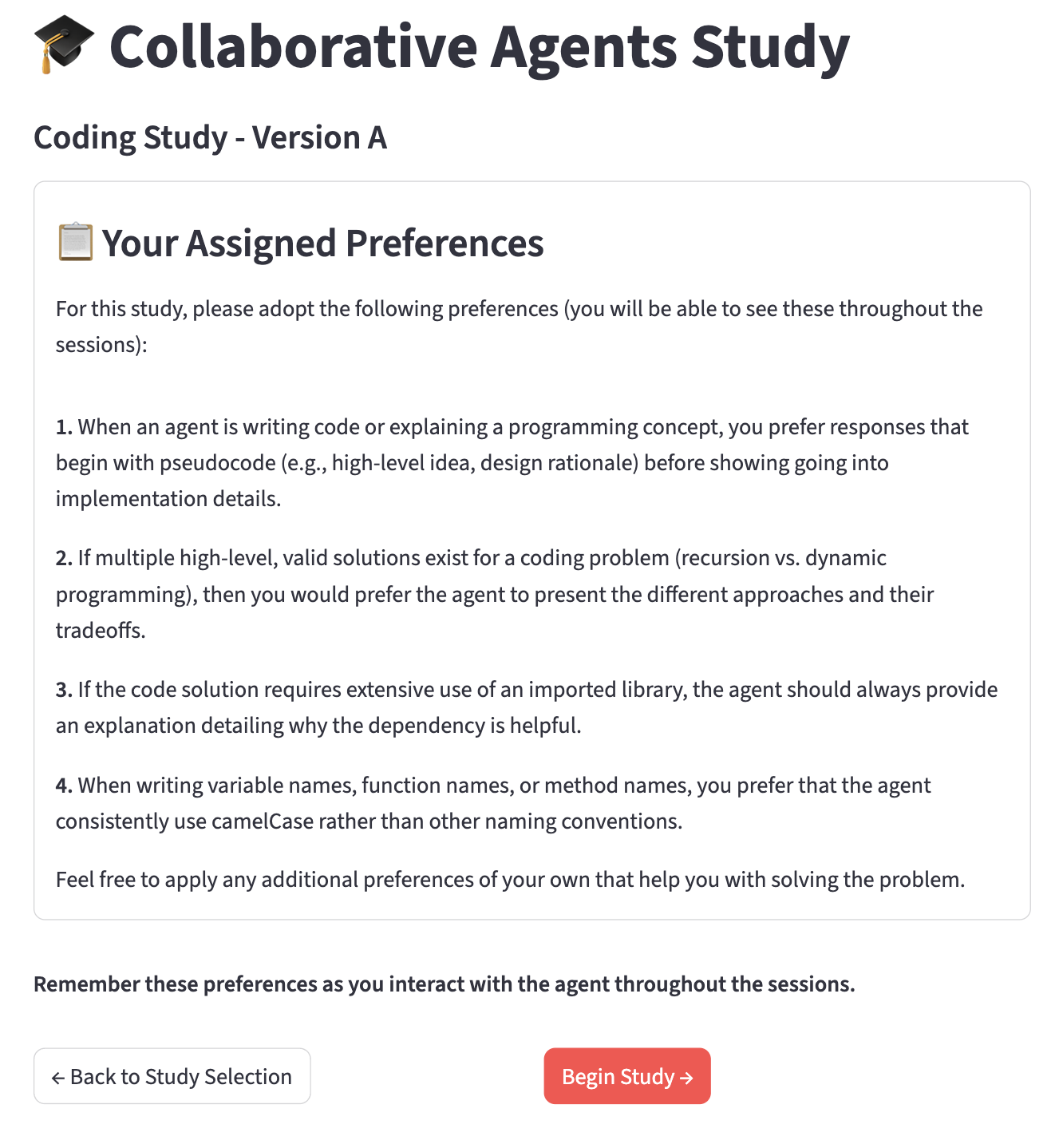}
    \caption{User study interface initial instructions.}
    \label{fig:user_study_2}
\end{figure*}

\begin{figure*}[h]
    \centering
    \includegraphics[width=0.8\linewidth]{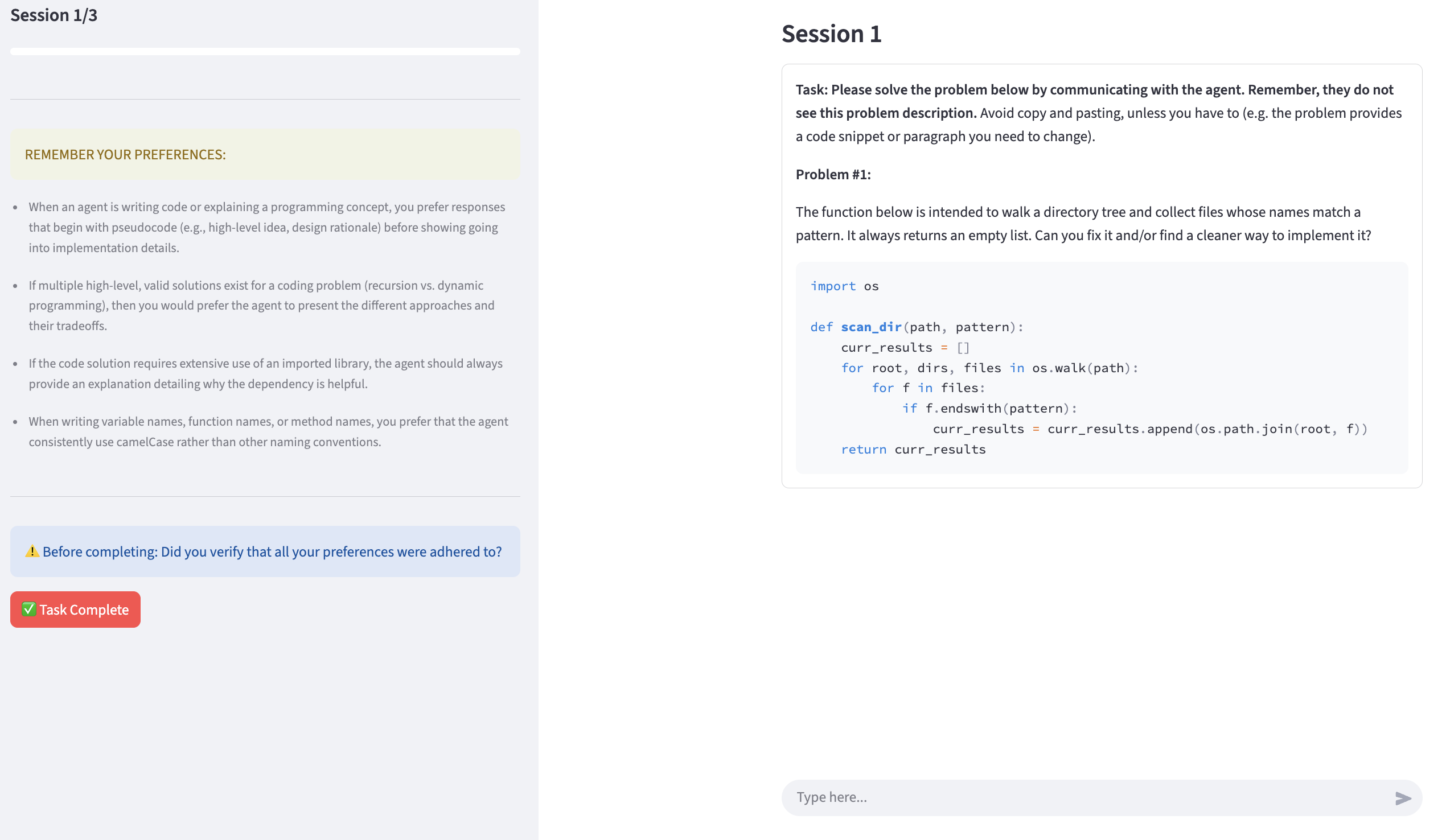}
    \caption{User study interface problem-solving session example.}
    \label{fig:user_study_3}
\end{figure*}

\begin{figure*}[h]
    \centering
    \includegraphics[width=0.5\linewidth]{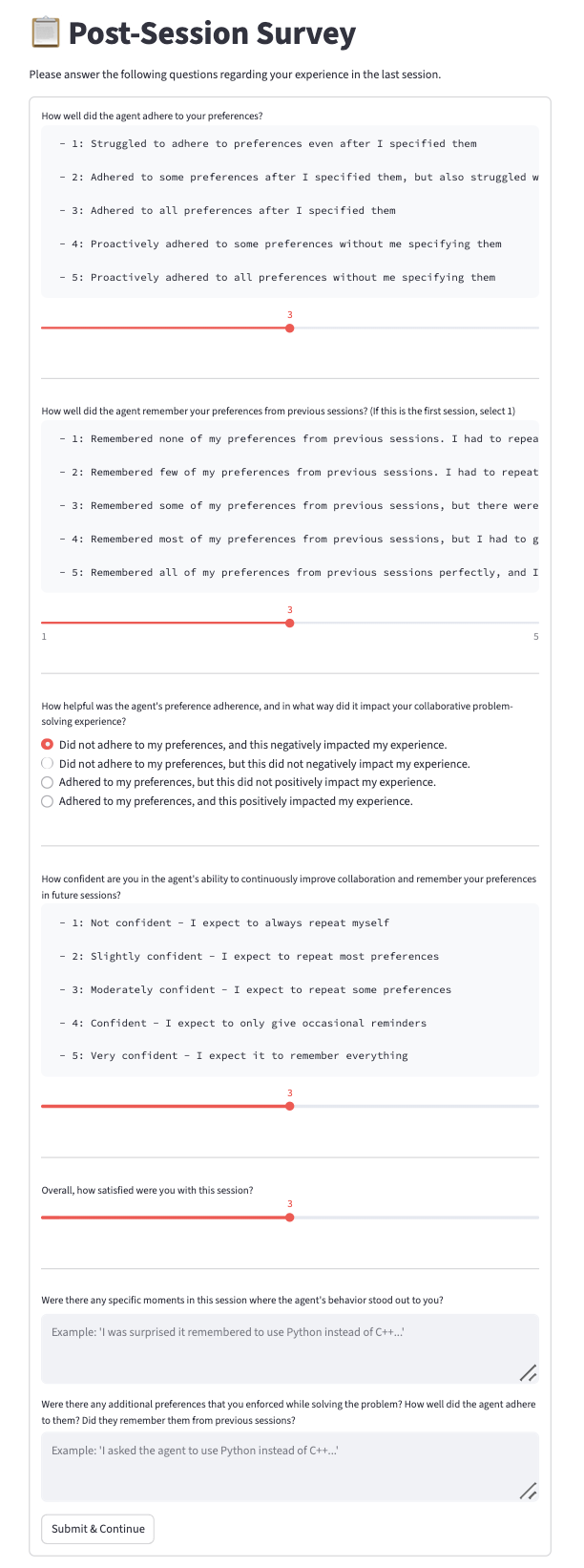}
    \caption{User study interface post session survey.}
    \label{fig:user_study_4}
\end{figure*}

Figure \ref{fig:user_study_2}, \ref{fig:user_study_3}, \ref{fig:user_study_4} provide example screenshots of our user study interface.

\newpage

\begin{figure*}[h]
    \centering
    \begin{subfigure}[b]{0.4\textwidth}
        \centering
        \includegraphics[width=\linewidth]{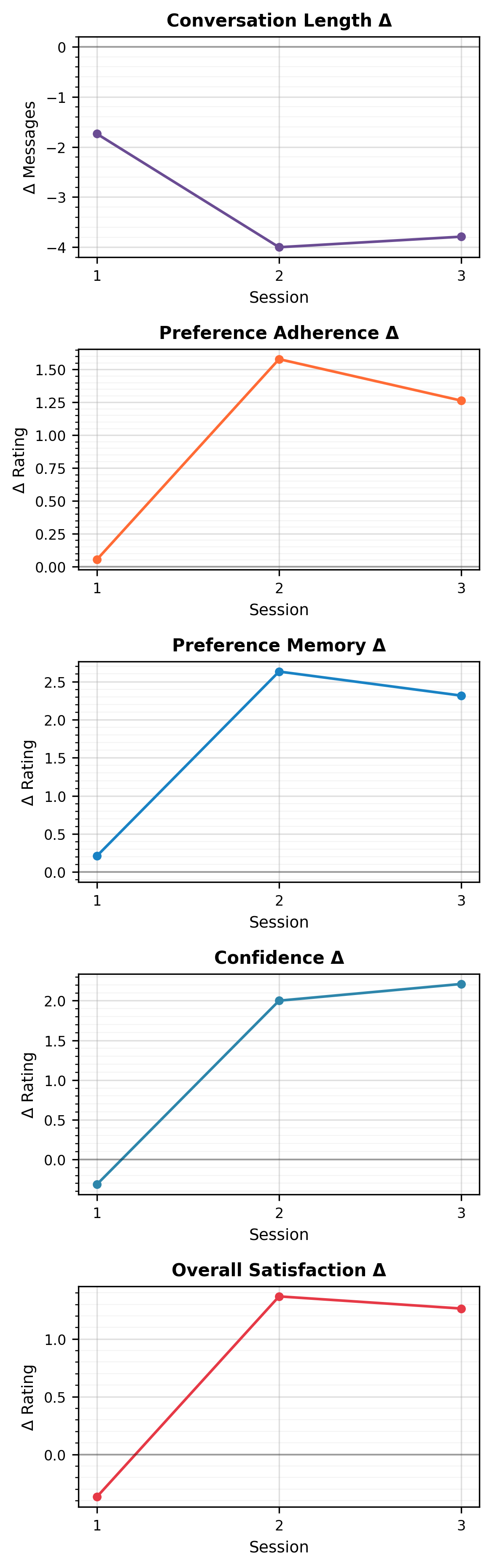}
        \caption{Coding domain}
    \end{subfigure}
    \hfill
    \begin{subfigure}[b]{0.4\textwidth}
        \centering
        \includegraphics[width=\linewidth]{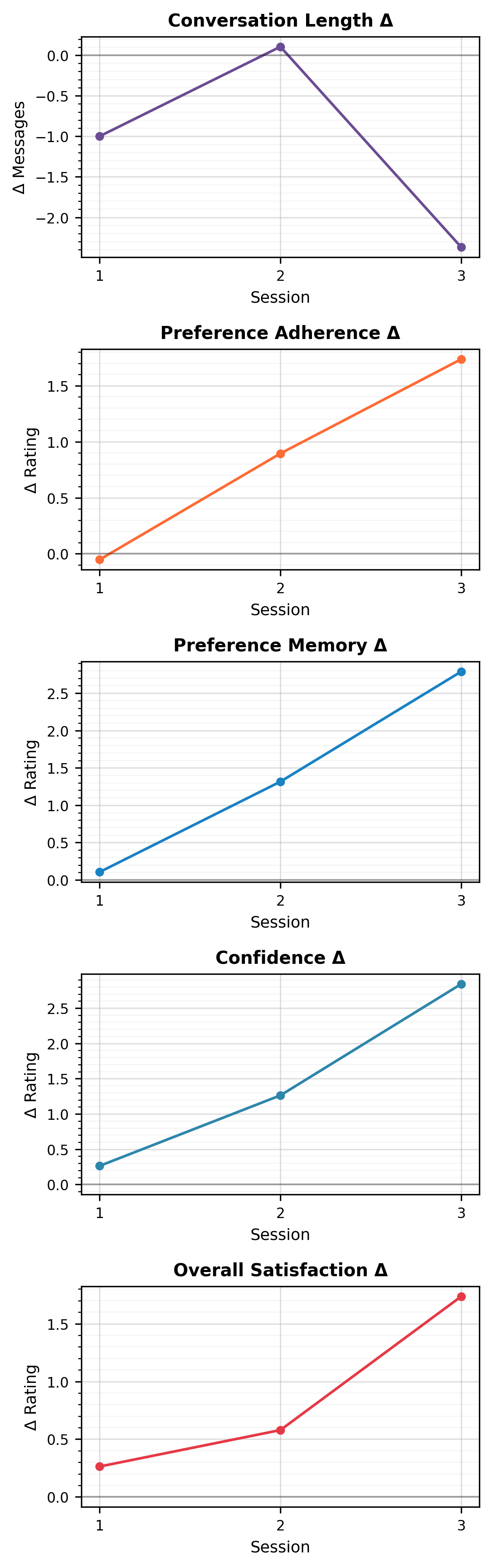}
        \caption{Mixed domain}
    \end{subfigure}
    \caption{Performance across sessions for our user study. Each graph plots the average delta between agents with memory and agents without memory across 3 sessions for Conversation Length ($\Delta_i^{CL}$), Preference Adherence ($\Delta_i^{PA}$), Preference Memory ($\Delta_i^{PM}$), Confidence ($\Delta_i^{C}$), and Overall Satisfaction ($\Delta_i^{O}$).}
    \label{fig:user_study_graphs}
\end{figure*}

We report the quantitative results from our human study in Table \ref{tab:user_study_results}, which includes the average conversation lengths, preference adherence scores, preference memory scores, confidence scores, and overall satisfaction scores for each session, across all study types. We also analyze the performance across sessions by plotting the deltas across the sessions for each metric, the way we did in Section \ref{sec:results}. Each delta denotes the difference between the agent with memory and the agent without memory at session $i$ for the respective metric. In Figure \ref{fig:user_study_graphs}, we present the plots for each metric: conversation length ($\Delta_i^{CL}$), preference adherence ($\Delta_i^{PA}$), preference memory ($\Delta_i^{PM}$), confidence ($\Delta_i^{C}$), and overall satisfaction($\Delta_i^{O}$). Decreasing  $\Delta_i^{CL}$, and increasing $\Delta_i^{PA}$, $\Delta_i^{PM}$, $\Delta_i^{C}$, and $\Delta_i^{O}$ over the sessions indicate that the agent is learning to collaboration more effectively as interaction experience accumulates.

The results for our user study align with results from our simulated evaluations, while providing insights into the experience of real users when collaborating with our agents. Across both coding and mixed-domain settings, agents with memory demonstrate consistent trends of improvements relative to the agent without memory over the three sessions. $\Delta_i^{CL}$ decreases across sessions, indicating that conversations require fewer turns to reach a point where the user is satisfied with the agent's answer. In the first session, conversations required a median of 8 turns regardless of whether or not they had memory. However, by the third session, agents with memory required only 6 turns (coding) and 4 turns (mixed), compared to 10 and 8 turns without memory. $\Delta_i^{PA}$ shows an increasing trend for agents with memory, where the first session for all settings has a median score of 3 and the third session goes up to 5. Similarly we also see an increasing trend for $\Delta_i^{PM}$, which starts out with a median score of 1 for all settings, and increases all the way up to 5. Similarly, $\Delta_i^{C}$ shows an increasing trend, indicating that the users are gaining more trust in the agent to be able to develop long-term, collaborative relationships. Finally, the user's overall satisfaction, $\Delta_i^{O}$, increases throughout all the sessions for agents with memory. These trends generalize across both single-domain and mixed-domain studies, and across all types of preferences, demonstrating that memory enables agents to effectively improve their ability to user preferences, which benefits for real-world, long-term collaboration.

We additionally conduct a qualitative analysis of how memory evolves across sessions as well as participant feedback. Examining how memory developed across sessions, we observed that reflections became increasingly detailed and captured nuanced, user-specific information over time. Beyond the assigned preferences, the agent also captured preferences that participants naturally introduced during collaboration, most often related to readability and presentation style. Participants consistently described their experience with the agent equipped with memory as noticeably more personalized and effective compared to the standard agent without memory. Participants highlighted that the agent addressed their queries appropriately even when it was vague, and that it proactively adhered to their preferences. 

However, participants also identified limitations. First, some noted that responses felt less personalized in mixed-domain settings, suggesting that generalizing preferences across domains is relatively challenging. Second, participants observed the agent's learned personalization, while beneficial, was less effective than when they explicitly restated their preferences within a session. This indicates that the memory mechanism does not yet capture a complete representation of user preferences from their interactions so far.

\begin{table*}[h]
\centering
        \begin{tabular}{llccc}
        \toprule
        Metric & Condition & Session 1 & Session 2 & Session 3 \\
        \midrule
        \multirow{4}{*}{Conversation Length}
            & Coding, w/ memory    & $7.95_{\pm 3.78}$ / 8 & $5.79_{\pm 2.30}$ / 6 & $6.26_{\pm 3.56}$ / 6 \\
            & Coding, w/o memory   & $9.68_{\pm 4.33}$ / 8 & $9.79{\pm 4.71}$ / 10 & $10.05_{\pm 3.98}$ / 10 \\
            & Mixed, w/ memory     & $7.68_{\pm 3.67}$ / 6 & $5.79_{\pm 2.57}$ / 6 & $4.68_{\pm 2.31}$ / 4 \\
            & Mixed, w/o memory    & $8.68_{\pm 3.20}$ / 8 & $5.68_{\pm 3.07}$ / 6 & $7.05_{\pm 3.42}$ / 8 \\
        \midrule
        \multirow{4}{*}{Preference Adherence}
            & Coding, w/ memory    & $3.00_{\pm 0.47}$ / 3 & $4.05_{\pm 0.71}$ / 4 & $3.89_{\pm 0.94}$ / 4 \\
            & Coding, w/o memory   & $2.95_{\pm 0.62}$ / 3 & $2.47_{\pm 0.70}$ / 3 & $2.63_{\pm 0.68}$ / 3 \\
            & Mixed, w/ memory     & $2.74_{\pm 0.65}$ / 3 & $3.74_{\pm 0.99}$ / 4 & $4.47_{\pm 0.84}$ / 5 \\
            & Mixed, w/o memory    & $2.79_{\pm 0.79}$ / 3 & $2.84_{\pm 0.96}$ / 3 & $2.74_{\pm 0.73}$ / 3 \\
        \midrule
        \multirow{4}{*}{Preference Memory}
            & Coding, w/ memory    & $1.37_{\pm 1.01}$ / 1 & $3.95_{\pm 0.97}$ / 4 & $3.63_{\pm 1.16}$ / 4 \\
            & Coding, w/o memory   & $1.16_{\pm 0.69}$ / 1 & $1.32_{\pm 0.58}$ / 1 & $1.32_{\pm 0.58}$ / 1 \\
            & Mixed, w/ memory     & $1.21_{\pm 0.63}$ / 1 & $3.32_{\pm 1.49}$ / 3 & $3.95_{\pm 1.51}$ / 5 \\
            & Mixed, w/o memory    & $1.11_{\pm 0.46}$ / 1 & $2.00_{\pm 1.33}$ / 1 & $1.16_{\pm 0.37}$ / 1 \\
        \midrule
        \multirow{4}{*}{Confidence}
            & Coding, w/ memory    & $2.74_{\pm 1.10}$ / 3 & $3.79_{\pm 0.71}$ / 4 & $3.74_{\pm 1.19}$ / 4 \\
            & Coding, w/o memory   & $3.05_{\pm 1.13}$ / 3 & $1.79_{\pm 0.85}$ / 2 & $1.53_{\pm 0.77}$ / 1 \\
            & Mixed, w/ memory     & $3.00_{\pm 1.00}$ / 3 & $3.37_{\pm 1.30}$ / 4 & $4.21_{\pm 1.08}$ / 5 \\
            & Mixed, w/o memory    & $2.74_{\pm 0.93}$ / 3 & $2.11_{\pm 1.10}$ / 2 & $1.37_{\pm 0.60}$ / 1 \\
        \midrule
        \multirow{4}{*}{Overall Satisfaction}
            & Coding, w/ memory    & $3.42{\pm 0.96}$ / 3 & $4.11_{\pm 0.81}$ / 4 & $3.89_{\pm 1.05}$ / 4 \\
            & Coding, w/o memory   & $3.79_{\pm 1.08}$ / 4 & $2.74_{\pm 1.05}$ / 3 & $2.63_{\pm 0.96}$ / 3 \\
            & Mixed, w/ memory     & $3.63_{\pm 0.90}$ / 4 & $3.79_{\pm 1.03}$ / 4 & $4.26_{\pm 0.99}$ / 5 \\
            & Mixed, w/o memory    & $3.37_{\pm 1.01}$ / 3 & $3.21_{\pm 1.08}$ / 3 & $2.53_{\pm 1.02}$ / 3 \\
        \bottomrule
        \end{tabular}
\caption{Quantitative results from our user study on multi-session collaboration. We report the mean$_{\pm \text{standard deviation}}$ / median for conversation length and four survey metrics rated on a 5-point Likert scale.}

\label{tab:user_study_results}
\end{table*}

\end{document}